\pdfoutput=1

\documentclass[journal]{IEEEtran}
\ifCLASSINFOpdf
\else
\fi
\usepackage{times}  
\usepackage[hyphens]{url}  
\usepackage{graphicx} 
\usepackage{caption} 
\usepackage{amssymb}
\usepackage{amsthm}
\usepackage{amsmath}
\usepackage[ruled,linesnumbered]{algorithm2e}
\usepackage{booktabs}
\usepackage{multirow}
\usepackage{subfigure}
\usepackage{microtype}
\usepackage{graphicx}
\usepackage{pgfplots}

\DeclareMathOperator*{\argmin}{arg\,min}

\usepackage{xspace}
\makeatletter
\DeclareRobustCommand\onedot{\futurelet\@let@token\@onedot}
\def\@onedot{\ifx\@let@token.\else.\null\fi\xspace}

\def\eg{\emph{e.g}\onedot} 
\def\ie{\emph{i.e}\onedot} 
 
\def\etc{\emph{etc}\onedot} 
 
\def\etal{\emph{et al}\onedot}

\begin{document}
%
\title{Learning Transferable Parameters for Unsupervised Domain Adaptation}
%
%
%

\author{Zhongyi Han, Haoliang Sun, Yilong Yin$^{*}$
\IEEEcompsocitemizethanks{\IEEEcompsocthanksitem Z. Han, H. Sun, and Y. Yin are with School of Software, Shandong University, Jinan 250101, China.\protect\\
E-mail: ylyin@sdu.edu.cn (Y. Yin)}
\thanks{Manuscript received XX, 21XX; revised XX, 21XX.}}

%
%

\markboth{IEEE Transactions on Image Processing,~Vol.~xx, No.~x, xx~21xx}
{Shell \MakeLowercase{\textit{et al.}}: Bare Demo of IEEEtran.cls for IEEE Journals}
%



\maketitle

\begin{abstract}

    Unsupervised domain adaptation (UDA) enables a learning machine to adapt from a labeled source domain to an unlabeled domain under the distribution shift. Thanks to the strong representation ability of deep neural networks, recent remarkable achievements in UDA resort to learning domain-invariant features. Intuitively, the hope is that a good feature representation, together with the hypothesis learned from the source domain, can generalize well to the target domain. However, the learning processes of domain-invariant features and source hypothesis inevitably involve domain-specific information that would degrade the generalizability of UDA models on the target domain. In this paper, motivated by the lottery ticket hypothesis that only partial parameters are essential for generalization, we find that only partial parameters are essential for learning domain-invariant information and generalizing well in UDA. Such parameters are termed transferable parameters. In contrast, the other parameters tend to fit domain-specific details and often fail to generalize, which we term as untransferable parameters. Driven by this insight, we propose Transferable Parameter Learning (TransPar) to reduce the side effect brought by domain-specific information in the learning process and thus enhance the memorization of domain-invariant information. Specifically, according to the distribution discrepancy degree, we divide all parameters into transferable and untransferable ones in each training iteration. We then perform separate updates rules for the two types of parameters. Extensive experiments on image classification and regression tasks (keypoint detection) show that TransPar outperforms prior arts by non-trivial margins. Moreover, experiments demonstrate that TransPar can be integrated into the most popular deep UDA networks and be easily extended to handle any data distribution shift scenarios.

\end{abstract}

\begin{IEEEkeywords}
Unsupervised domain adaptation, transferable parameter learning, image classification, keypoint detection.

\end{IEEEkeywords}

%
\IEEEpeerreviewmaketitle

\section{Introduction}

\IEEEPARstart{U}{nsupervised} domain adaptation (UDA) has proven highly successful in handling distribution shift and knowledge transfer. On the one hand, conventional supervised learning algorithms perform well under static environments where we draw training and test examples from an identical data distribution. If the training and test distributions are substantially different, the conventional supervised learning algorithms often fail to generalize~\cite{valiant1984theory}. In practice, we expect our learning machines to generalize well on the test domain of interest that is similar yet distinct from the training domain. UDA is the key machine learning topic to deal with this dilemma~\cite{ben2007analysis}. On the other hand, deep neural networks (DNNs) are applied to an increasingly broad array of applications but at the expense of laborious large-scale training data annotation. To avoid expensive data labeling, UDA takes on the critical responsibilities of transferring knowledge from previously labeled datasets in a transductive manner~\cite{DBLP:journals/pami/ManciniPBCR21}. 

Prominent theoretical advances in UDA emerge a consensus that the generalization error of the source hypothesis (\eg, classifier, regressor) with respect to the target domain is upper bounded by the empirical source risk, the domain dissimilarity, the ideal joint hypothesis error, and a factor consisting of the hypothesis complexity and sample size~\cite{ben2007analysis, DBLP:journals/pami/KouwL21}. Unlike conventional supervised learning bounds, the measurement of domain dissimilarity is the key to bridge the source and target domains. Currently, the most popular measures of domain dissimilarity include the $\mathcal{H} \Delta \mathcal{H}$ distance~\cite{ben2010theory}, maximum mean discrepancy ~\cite{DBLP:conf/nips/GrettonBRSS06}, discrepancy distance~\cite{mansour2009domain}, and margin disparity discrepancy~\cite{DBLP:conf/icml/0002LLJ19}.

\begin{figure}[t]
  \centering
  \includegraphics[width=0.5\textwidth]{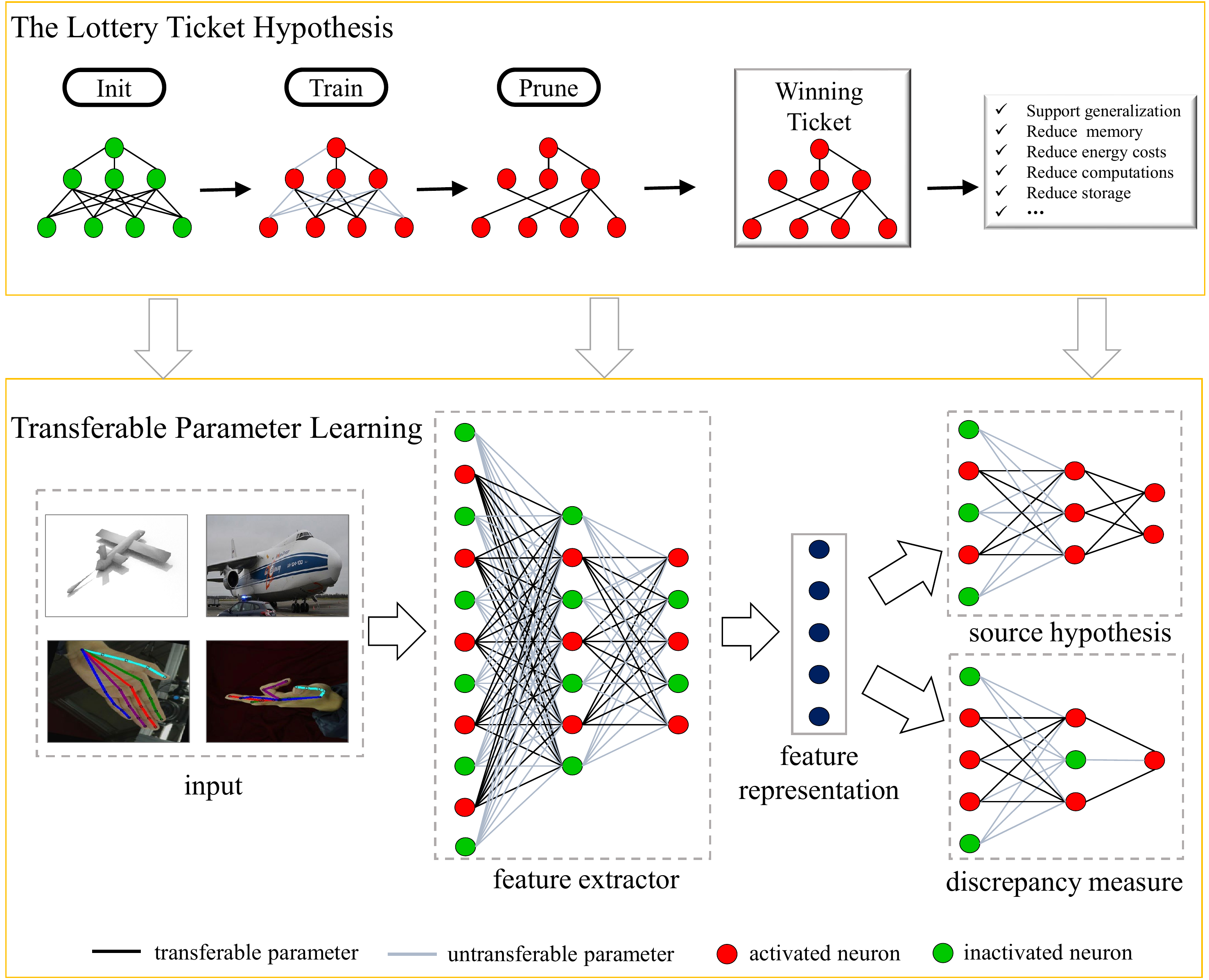}
  \caption{The lottery ticket hypothesis shows that DNNs include some winning tickets (parameters) that support generalization. Motivated by it, we find that deep UDA networks include partial parameters that are essential for generalization on the target domain. Accordingly, we propose transferable parameter learning to divide the parameters into transferable parameters and untransferable parameters. The input images stand for the image classification and keypoint detection tasks.}
  \label{TransPar}
\end{figure}


On par with the above important theoretical findings, deep UDA methods resort to learning the domain-invariant feature representation to minimize the distribution discrepancy. Great studies in deep UDA can be divided into two categories: \textit{1) moment matching.} The core idea is to minimize the discrepancy of high-dimensional features cross domains~\cite{pan2010domain,long2017deep}; \textit{2) adversarial training.} The core idea is to learn transferable feature representations by confusing a domain discriminator in a two-player game, yielding the state-of-the-art performance on many visual tasks~\cite{DBLP:conf/icml/GaninL15,tzeng2017adversarial,DBLP:conf/nips/LongC0J18}. Intuitively, they expect a good feature representation and a hypothesis learned from the source domain to generalize well to the target domain. However, during learning domain-invariant features and source hypotheses, much unnecessary domain-specific information is still involved, thus compromising the generalization to the target domain. Therefore, if we can reduce the side effect of domain-specific information before the testing time, the generalizability and robustness of deep UDA will improve.

In essence, the parameters in deep UDA networks dominant the learning of transferable feature representation and source hypotheses. Recent interesting studies have theoretically proven that over-parameterization is one of the key factors that has contributed to the successful training of DNNs, but allowing them to overfit the domain-specific information of the source domain easily~\cite{DBLP:journals/tit/SoltanolkotabiJ19, pmlr-v97-allen-zhu19a}. The \textit{lottery ticket hypothesis}~\cite{DBLP:conf/iclr/FrankleC19} shows that only partial parameters are important for generalization (see Fig.~\ref{TransPar}). Motivated by that, it remains a question whether we can divide the parameters of any deep UDA network into two parts to reduce the side effect of domain-specific information. In this way, we can enhance the learning of domain-invariant features and source hypotheses and further improve deep UDA networks' generalization performances on the target domain.

In this paper, we present a novel insight: \textit{rather than learning the domain-invariant feature representation, it is better to learn domain-invariant parameters.} Interestingly, as illustrated in Fig.~\ref{TransPar}, we find that only partial parameters are essential for learning domain-invariant information and generalizing well in UDA, which we term as \textit{transferable parameters}. In contrast, the other parameters tend to fit domain-specific information and usually fail to generalize, which we term as \textit{untransferable parameters}. Driven by this insight, we propose Transferable Parameter Learning (TransPar) to reduce the side effect of domain-specific information in the learning process and thus enhance the learning of domain-invariant information. Specifically, TransPar first identifies the transferable parameters among all parameters of a deep UDA network by exploiting the largest-magnitude weights and distribution distance in each iteration. In this way, TransPar categorizes all parameters into two parts, \ie, the transferable and untransferable parameters. TransPar then suggests separate update rules for these two types of parameters. For the transferable parameters, TransPar performs the robust positive update. This part of the parameters is updated using the gradients derived from the objective function and weight decay. For the untransferable parameters, TransPar performs the negative update. Their values are penalized with the weight decay and without the gradients derived from the objective functions.

As the first effort to identify the transferable parameters in UDA, two critical questions emerge and determine the success of domain adaptation: 

\textit{Firstly, how to set the ratio between transferable and untransferable parameters?} Our answer is according to the distribution distance. Intuitively, the smaller the distribution distance between source and target domains, the larger the ratio of the transferable parameters because the domain-specific information is minor. However, the variational distance of real-valued distributions cannot be measured from finite samples~\cite{DBLP:conf/focs/BatuFRSW00}. Instead, we propose to estimate the ratio between transferable and untransferable parameters by using logistic regression classifiers to measure the proxy $\mathcal{A}$-distance~\cite{ben2007analysis}. 

\textit{Secondly, which module parameters are dividable as transferable and untransferable parameters in deep UDA networks?} Most deep UDA networks comprise three modules: a feature extractor to learning invariant representation, a source hypothesis to optimize the objective function based on the labeled source data, and a domain discrepancy measurement module (\eg, domain discriminator) to minimize the distribution distance of high-level features. We find that 1) the parameters of the feature extractor and the domain discrepancy measurement are dividable because they promote each other; 2) the parameters of the source hypothesis can be dividable only if the objective function adds a term about the unsupervised loss of target domain (\ie, entropy minimization loss).

We summarize our contributions as follows:
\begin{itemize} 
    \item To the best of our knowledge, it is the first time that finds only partial parameters are important for learning domain-invariant information and generalizing well in UDA. This exciting insight could open a new door to address the problem of distribution shift and knowledge transfer.
    \item We propose a novel and effective method to categorize the transferable and untransferable parameters according to the largest-magnitude weights and the distribution distance. The proposal is suitable to any deep UDA network and can be easily extended to handle any case of data distribution mismatches (\eg, covariate shift, target shift).
    \item We empirically validate the proposal on various UDA datasets and settings, on which it achieves remarkable improvements compared with the original deep UDA networks. For example, after integrated our proposal into domain adversarial neural network (DANN)~\cite{DBLP:conf/icml/GaninL15}, the average accuracy is improved from 82.2\% to 87.8\% on the Office-31 dataset.
\end{itemize}

In the following, we first briefly review related work in Section~\ref{related_work}. Then, we fulfill the learning set-up and short introduction for unsupervised domain adaptation networks in Section~\ref{statement}. Next, we present a detailed description of our proposed methods in Section~\ref{Method}. Experimental results on the image classification and keypoint detection tasks are in Section~\ref{Experiment}. Finally, we close this paper in Section~\ref{Conclusion}.

\section{Related Work} 
\label{related_work}

UDA is gaining momentum in the past decade as an effective strategy to cope with the distribution shift cross domains. Previous works have fulfilled in-depth research on theoretical and methodological aspects, which have firmly established the research foundation of this topic.


From the view of the distribution shift hypothesis, the existing theoretical works can be broadly categorized into the covariate shift assumption-based works and the target shift assumption-based works. (\romannumeral1) The covariate shift assumption-based works suppose the marginal distributions of features cross domains differ, and all other conditions are consistent. The main idea lies in the distribution distances compared to conventional supervised learning. One of the pioneering distribution distances in this line was proposed by Ben-David \etal~\cite{ben2007analysis} who proposed the first distribution distance termed $\mathcal{H} \Delta \mathcal{H}$ distance that defeats the difficulties of empirical risk estimation with limited samples. Motivated by $\mathcal{H} \Delta \mathcal{H}$ distance, Mansour \etal~\cite{mansour2009domain} proposed the novel distance, discrepancy distance, that is tailored to generalized loss functions, and derived the data-dependent Rademacher complexity learning bounds. Zhang \etal~\cite{DBLP:conf/icml/0002LLJ19} proposed the margin disparity discrepancy and derived the rigorous margin-based generalization bounds to multi-classification domain adaptation tasks for the first time. (\romannumeral2) The target shift assumption-based works assume that the source and target label distributions probably substantially differ. Representative works include: Zhang \etal~\cite{zhang_domain_2013} investigated this assumption from a causal inference perspective; Germain \etal~\cite{germain_pac-bayesian_2013} introduced the first PAC-Bayesian theoretical analysis for this assumption; In this context, Courty \etal~\cite{courty_joint_2017} proposed the joint distribution optimal transportation theory. Note that our method is suitable for both types of assumptions.

On par with these theoretical findings, there is rich progress in UDA algorithms. Previous works can be assorted into the importance estimation, the moment matching, the pseudo labeling, and the adversarial training. (\romannumeral1) The philosophy behind the importance estimation is to find distribution intersection cross domains. The KLIEP algorithm proposed by Sugiyama \etal~\cite{sugiyama_direct_2008} is the representative work. (\romannumeral2) The core idea behind the moment matching is to minimize the discrepancy of high-dimensional statistics cross domains. Some representative works include maximum mean discrepancy~\cite{huang_correcting_2006}, transfer component analysis~\cite{pan2010domain}, and deep adaptation networks~\cite{long2017deep,long_learning_2015}.  (\romannumeral3) The main idea behind the pseudo labeling is attempting to assign pseudo labels to target domain samples and selecting reliable labels to complete supervised training. Some characteristic works involve transduction with domain shift algorihtm~\cite{sener2016learning}, asymmetric tri-training~\cite{saito2017asymmetric}, and collaborative and adversarial network~\cite{zhang_collaborative_2018}. (\romannumeral4) Another route of research studies on the interesting adversarial training and introduces a domain discriminator to distinguish the samples between two domains for learning the domain-invariant representation~\cite{saito_maximum_2018,DBLP:conf/nips/LongC0J18}. One of the pioneering theoretical works in this line was conducted by Ganin \etal~\cite{DBLP:conf/icml/GaninL15} who proposed the first deep adversarial adaptation network.

Our study clearly differs from current deep UDA approaches in two aspects. First, this study attempts to learn invariant parameters rather than learning invariant representations so that we reach the essence of the distribution shift problem. Second, our proposal is so general and flexible that it can be integrated into any deep UDA network, \ie, our proposal is compatible with any of these networks with a simple modification.

Existing UDA algorithms have significant application values in image classification, semantic segmentation, regression (\eg, 2D keypoint detection), \etc. 2D keypoint detection is an innovative task that has become a hot topic in recent years due to its widespread utility in computer vision applications. Representative works include single round message passing networks~\cite{DBLP:conf/nips/TompsonJLB14}, stacked hourglass networks~\cite{DBLP:conf/eccv/NewellYD16}, deconvolutional layers based ResNet~\cite{DBLP:conf/eccv/XiaoWW18}, and high-resolution networks~\cite{DBLP:conf/cvpr/0009XLW19}. Notably, Jiang \etal~\cite{DBLP:journals/corr/abs-2103-06175} proposed the regressive domain adaptation (RegDA) for unsupervised 2D keypoint detection. \textit{Note that our method is not intended to refine the network architecture further but to validate the effectiveness of transferable parameter learning of domain adaptation in 2D keypoint detection.}


The lottery ticket hypothesis~\cite{DBLP:conf/iclr/FrankleC19} confirms that overparameterized DNNs contain winning tickets (parameters) that are important for generalization. With this part of winning tickets, one can train the tiny and sparsified networks to generalize well and exceed the test accuracy of the original DNNs. Although the lottery ticket hypothesis inspires this study, this study is essentially different from it. The lottery ticket hypothesis aims to achieve network compression. It focuses on gaining a sparsified subnetwork that has competitive generalization performance compared with the original network. This paper focuses on the UDA problem. We aim to find the transferable and untransferable parameters to reduce the side effect of domain-specific information, which significantly advances the deep UDA network's generalization performance on target domains. Recent few interesting works have validated the effectiveness of lottery ticket hypothesis on learning with noisy labels~\cite{DBLP:conf/iclr/XiaL00WGC21, DBLP:journals/corr/abs-2102-11628}, and out-of-distribution generalizaiton~\cite{DBLP:journals/corr/abs-2106-02890}. \textit{Different from these extensions, we, for the first time, verify that the learning of transferable parameters is feasible in UDA, propose a novel transferable/untransferable ratio estimation method, and investigate which module parameters are dividable to find winning tickets in UDA networks.}

\section{Deep UDA Statement}
\label{statement}

In this section, we first introduce the necessary notations for learning transferable parameters in UDA. Then, we present common structures of deep UDA networks and their optimization processes.

\subsection{Learning Set-Up}
In UDA, we denote by $\mathbf{x}$ the input, $\mathbf{y}$ the label, and $d$ the Bernoulli variable indicating to which domain $\mathbf{x}$ belongs. In our terminology, a sample of $n_s$ labeled training examples $\{(x_s^i, y_s^i)\}_{i=1}^{n_s}$ ($d=1$) is drawn according to the source distribution $p$ defined on $\mathcal{X} \times \mathcal{Y}$. Meanwhile, a sample of $n_t$ unlabeled test examples $\{(x_t^i)\}_{i=1}^{n_t}$ ($d=0$) is drawn according to the target distribution $q$ that somewhat differs from $p$. $\mathcal{X}$ is the input set and $\mathcal{Y}$ is the label set. $\mathcal{Y}$ is $\{-1, +1\}$ in binary classification, $\{1, \dots, K\}$ in multi-class classification, or $\mathcal{Y}\in \mathbb{R}^2$ in regression problem (\eg, the keypoint detection task). Assume that the distributions $p(\mathbf{x})$ and $q(\mathbf{x})$ should not be dissimilar substantially~\cite{DBLP:journals/jmlr/Ben-DavidLLP10}. We denote by $f: \mathcal{X} \rightarrow \mathcal{Y}$ a deep UDA model. Our objective is to learn a good deep UDA model to generalize well on the test domain. We denote vectors and matrices by bold-faced letters. We denote the  inner product between two vectors by $\langle\cdot,\cdot\rangle$. We use $\|\cdot\|_p$ as the $\ell_p$ norm of vectors or matrices. For a function $f$, we use $\triangledown f$ to denote its gradient. Let $[n]=\{1,2,\dots,n\}$. We denote by $\mathcal{L}: \mathcal{Y} \times \mathcal{Y} \rightarrow \mathbb{R}$ a loss function defined over pairs of labels.

\subsection{Deep UDA Networks and Optimization Process}
Recall that prominent theoretical advances in UDA consent that the expected target error is upper bounded by three terms: the expected source risk, the expected domain discrepancy, and the idea joint hypothesis error. Formually, assume the objective function $\mathcal{L}$ is symmetric and obeys the triangle inequality. Then, for any hypothesis, $h \in \mathcal{H}$, the expected target risk $\epsilon_{q}(h)$ is bounded by the following inequality:

\begin{equation}
    \epsilon_{q}(h) \le \epsilon_{p}(h) + |\epsilon_{p}(h, h^*) - \epsilon_{q}(h, h^*)| + \lambda \,,
    \label{error bound}
\end{equation}
where the first term on the right-hand side refers to expected source risk, the second term represents the measurement of distribution discrepancy, and $\lambda \!=\! \epsilon_{Q_n}(h^*)\! + \!\epsilon_{P}(h^*)$ is the ideal combined error of optimal hypothesis $h^*$ that achieves the minimum risk on the source and the target domains:
\begin{equation}
    h^* = \argmin_{h\in \mathcal{H}} \epsilon_p + \epsilon_q(h)\,.
\end{equation}

Since the ideal joint hypothesis error $\lambda$ is assumed to be small enough~\cite{DBLP:journals/jmlr/Ben-DavidLLP10}, most deep UDA networks focus on the expected source risk and the distribution discrepancy. Without losing generality, deep UDA networks mainly comprise three modules: a feature extractor $F$ parameterized by $\mathbf{\mathcal{W}}_f$ to learning invariant representation, a source hypothesis $C$ parameterized by $\mathbf{\mathcal{W}}_c$ to optimize the empirical source risk based on the labeled source data, and a domain discrepancy measurement module $D$ parameterized by $\mathbf{\mathcal{W}}_d$ to minimize the distribution distance of high-level features produced by the feature extractor. While different methods have different domain discrepancy measurement modules, most of them have an extra classifier parameterized by $\mathbf{\mathcal{W}}_d$ whatever they are moment matching, pseudo labeling, or adversarial training based algorithms. Note that our method is still suitable for special deep UDA networks (\eg, adversarial discriminative domain adaptation (ADDA)) that do not have the domain discrepancy measurement module.

The optimization goal of deep UDA can be stated as 

\begin{equation}
    \min \mathcal{L}(\mathbf{\mathcal{W}}_f, \mathbf{\mathcal{W}}_c, \mathbf{\mathcal{W}}_d) =  \min [\mathcal{L}^s(\mathbf{\mathcal{W}}_f, \mathbf{\mathcal{W}}_c) - \mathcal{L}^d(\mathbf{\mathcal{W}}_f, \mathbf{\mathcal{W}}_d)]\,,
\end{equation}
where 
\begin{equation}
    \mathcal{L}^s(\mathbf{\mathcal{W}}_f, \mathbf{\mathcal{W}}_c) = \mathcal{L}^s_{\mathbf{x} \sim p(\mathbf{x})}(\mathbf{\mathcal{W}}_f, \mathbf{\mathcal{W}}_c; (C\circ F(\mathbf{x}), \mathbf{y}))
\end{equation}
is the objective function based on the labeled source data, where $\circ$ denotes function composition. $\mathcal{L}^s$ is usually a cross-entropy loss function or a square loss function. Moreover,

\begin{equation}
    \mathcal{L}^d(\mathbf{\mathcal{W}}_f, \mathbf{\mathcal{W}}_d) = \mathcal{L}^d_{\mathbf{x} \sim p(\mathbf{x}) \, \text{or} \, q(\mathbf{x})}(\mathbf{\mathcal{W}}_f, \mathbf{\mathcal{W}}_d; (D\circ F(\mathbf{x}), \bullet))\,,
\end{equation}
is the domain discrepancy objective function based on the labeled source data and unlabeled target data. Here, the symbol $\bullet$ denotes the ground truth depending on specific domain discrepancy measurements. According to the most commonly used stochastic gradient descent algorithm (SGD), the update rules of the parameters $\mathbf{\mathcal{W}}_f, \mathbf{\mathcal{W}}_c, \mathbf{\mathcal{W}}_d$ can be represented by the following formulae, respectively:

\begin{align}
    \label{f}
    & \mathbf{\mathcal{W}}_f(t+1) \rightarrow \mathbf{\mathcal{W}}_f(t) - \eta\left(\frac{\partial (\mathcal{L}^s - \mathcal{L}^d)}{\partial \mathbf{\mathcal{W}}_f(t)} + \lambda \text{sgn}(\mathbf{\mathcal{W}}_f(t))\right)\,,\\ 
    \label{d}
    & \mathbf{\mathcal{W}}_d(t+1) \rightarrow \mathbf{\mathcal{W}}_d(t) - \eta\left(\frac{\partial (\mathcal{L}^d)}{\partial \mathbf{\mathcal{W}}_d(t)} + \lambda \text{sgn}(\mathbf{\mathcal{W}}_d(t))\right), \, \text{and}\\
    \label{c}
    & \mathbf{\mathcal{W}}_c(t+1) \rightarrow \mathbf{\mathcal{W}}_c(t) - \eta\left(\frac{\partial (\mathcal{L}^s)}{\partial \mathbf{\mathcal{W}}_c(t)} + \lambda \text{sgn}(\mathbf{\mathcal{W}}_c(t))\right)\,,
\end{align}
where $\mathbf{\mathcal{W}}_\ast(t)$ denotes the the set of parameters of a module~($\ast$) at the $t$-th iteration, $\eta > 0$ denotes the learning rate, and $\text{sgn}(\cdot)$ is the standard sgn function in mathematics. The term $\lambda \text{sgn}(\mathbf{\mathcal{W}}_c(t))$ denotes the weight decay technique that can decline the weights to smaller values, in which $\lambda$ is the weight decay coefficient. The weight decay performs similar to $\|\mathbf{\mathcal{W}}_\ast(t)\|_2$ normalization that can effectively mitigate the network over-fitting problem in the training process.

\section{Transferable Parameter Learning}
\label{Method}

In this section, we introduce Transferable Parameter Learning (TransPar) to answer our vital concerns: How to determine the transferable and untransferable parameters? How to judge the scope of transferable parameters? How to choose the ratio of transferable parameters? Moreover, how to update transferable and untransferable parameters?

\subsection{Identifying Transferable Parameters}
The objective of identifying transferable parameters is to divide the parameters of a module into a set of transferable parameters and a set of untransferable parameters. Inspired by the lottery ticket hypothesis, we think that only a set of essential parameters of a module in deep UDA networks contributes to learning domain-invariant information. Such parameters are called transferable parameters. In contrast, a set of parameters contributes to learning domain-specific information that hurt the generalization of deep UDA networks on the target domain, which we call untransferable parameters. Suppose we design an identifying criterion to select the set of transferable parameters precisely. In that case, we can reduce the side effect of domain-specific information in the learning process and thus enhance the generalization. To achieve that, we design an effective identifying criterion as follows.

General speaking, the designed identifying criterion in this paper exploits the largest-magnitude weights. Consider a parameter, denoted by $\text{w}_i(t) \in \mathbf{\mathcal{W}}_\ast(t)$, its gradient is $\triangledown \mathcal{L}(\text{w}_i(t))$ at the $t$-th iteration. The identifying criterion is defined by 

\begin{equation}
    \mathcal{T}_i(t) = |\triangledown \mathcal{L}(\text{w}_i(t)) \times \text{w}_i(t)|, i \in [m_\ast]\,,
    \label{criterion}
\end{equation}
where $m_\ast$ is the parameter number of a module in a deep UDA network. If the value of $\mathcal{T}_i(t)$ is large, $\text{w}_i(t)$ is viewed as a transferable parameter. On the contrary, if the value of $\mathcal{T}_i(t)$ is small, \eg, zero or very close to zero, $\text{w}_i(t)$ is regarded to be an untransferable parameter. It is not important for fitting domain-invariant information. If we update it, it will tend to fit domain-specific information.

As shown in Eq.~\eqref{criterion}, we design an iterative identifying manner that considers the activity of each parameter during the whole training process. This iterative manner is better than the one-shot manner~\cite{DBLP:conf/iclr/FrankleC19, DBLP:journals/corr/abs-2106-02890} that only considers the activity of each parameter at once iteration. When only considering the parameter activity at the first iteration, the activity difference between transferable and untransferable parameters is slight, \ie, the values of gradient and weight are indistinguishable. When only considering the parameter activity at the last iteration, some untransferable parameters already overfit the domain-specific information, leading to the invalid gradient information and the indistinguishable weight value. Unlike the one-shot manner, this iterative manner can observe the long-term activities of each parameter. The forward propagation of deep UDA networks depends on the magnitude of the weights, and the backward propagation depends on the gradients. Both weight and gradient interact in the forward and backward propagation to precisely estimate the importance of each parameter.  

We argue the significant advantages of exploiting the values of both gradient and weight from three aspects. First, the gradient and weight are complementary in different training stages. In the early stage, the values of weights are indistinguishable, whatever for the transferable and untransferable parameters. In contrast, the gradients of transferable parameters flow powerfully, indicating that the transferable parameters are activated. In the middle stage, the values of weights and their gradients are very influential. In the late state, the values of gradients drop to zero with the decrease of loss values, while the values of weights are significant. Second, when we only exploit gradient information, the parameter is identified as an untransferable parameter if the gradient value closes to zero. However, in the late stage, transferable parameters also have small gradient values and would be identified incorrectly, such that the generalization will be degraded. Finally, exploiting the values of both gradient and weight has been demonstrated effectively in the prominent work about label noise learning~\cite{DBLP:conf/iclr/XiaL00WGC21}.

\subsection{Judging the Scope of Transferable Parameters}
We respectively dissect the parameters of three standard modules of deep UDA networks as follows.

Firstly, it is reasonable to divide the parameters of the feature extractor into transferable parameters and untransferable parameters. The feature extractor module enables learning class-conditional domain-invariant features to confuse or encourage the domain discrepancy measurement and support the source hypothesis. Only partial parameters are essential for learning domain-invariant features, while partial parameters tend to fit domain-specific features. Driven by the domain discrepancy loss and the classification loss (Eq.~\eqref{f}), the parameters that are beneficial to learn class-conditional domain-invariant features have big gradient flows and significant values of weights. 

Secondly, it is also reasonable to divide the parameters of the domain discrepancy measurement module. In contrast to the feature extractor, driven by the adversarial objective function, the parameters in this module with big gradient flows and significant weights are essential to learning domain-specific information. If we only reserve these critical parameters, the domain discriminator becomes more discriminative.

Finally, the optimization goal of the source hypothesis based on the objective function (Eq.~\eqref{c}) is to achieve zero prediction error on the source data. Intuitively, the parameters of the source hypothesis tendency to fitting source information. It is uncertain whether the parameters of the source domain model contain transferable parameters or not. To ensure the source hypothesis has transferable parameters, we propose to add an unsupervised objective function to exploit the target information. Suppose the source hypothesis optimizes the supervised objective function based on the source data and the unsupervised objective function based on the target data. In that case, the source hypothesis fully considers both the information cross domains. The parameters that are beneficial to learn domain-invariant information have big gradient flows and significant values of weights. In such a way, it is reasonable to divide the parameters into transferable parameters and untransferable parameters. Specifically, we add common entropy loss to realize the entropy minimization of target data. The objective function of the source hypothesis is revised into

\begin{equation}
    \begin{split}
        \mathcal{L}^s(\mathbf{\mathcal{W}}_f, \mathbf{\mathcal{W}}_c) = &\mathcal{L}^s_{\mathbf{x} \sim p(\mathbf{x})}(\mathbf{\mathcal{W}}_f, \mathbf{\mathcal{W}}_c; (C\circ F(\mathbf{x}), \mathbf{y})) \\
        &+ \alpha \mathcal{L}^{ent}_{\mathbf{x} \sim q(\mathbf{x})}(\mathbf{\mathcal{W}}_f, \mathbf{\mathcal{W}}_c; C\circ F(\mathbf{x}))\,,
    \end{split}
\end{equation}
where $\mathcal{L}^{ent}$ denotes the common entropy loss function, and $\alpha$ is the coefficient that controls the impact of the entropy loss. 

In summary, the parameters of all three modules can be divided into transferable and untransferable parameters. Since the weights and gradients of the three models have different scales and ranges due to different objective functions, we consider their parameters separately to determine the transferable parameters in the next section.
 
\subsection{Determining the Ratio of Transferable Parameters}
We have presented how to identify the transferable parameters and judge the scope of transferable parameters. We exploit the distribution discrepancy degree to help divide the parameters into the transferable/untransferable ones. Intuitively, if the distribution discrepancy degree is high, the domain-invariant information is small. The number of required transferable parameters for learning domain-invariant information is then small. In other words, the number of transferable parameters has a negative correlation with distribution discrepancy degree. We therefore use the distribution discrepancy degree to help identify the ratio of transferable parameters. However, the variational distance of real-valued distributions cannot be measured from finite samples~\cite{DBLP:conf/focs/BatuFRSW00}. Instead, we propose to measure the proxy $\mathcal{A}$-distance~\cite{ben2007analysis} by the logistic regression classifier to compute the ratio between transferable and untransferable parameters. Given the empirical distributions $\hat{p}$ and $\hat{q}$, for a logistic regression classifier set $\mathcal{H}$, we denote by $d_\mathcal{A}$ the proxy $\mathcal{A}$-distance, which is defined by 

\begin{equation}
    d_\mathcal{A}(\hat{p}, \hat{q}) =  1 - 2 \min_{h\in \mathcal{H}}\text{err}(h)\,,
    \label{d_A}
\end{equation}
where $\text{err}(h)$ denotes the error of a two-class classifier $h$ on the task of discriminating between examples sampled from source or target distributions. In practice, the classifier is a standard domain discriminator that contains two fully connected layers. The inputs of the domain discriminator are the features extracted by the ImageNet-pretrained feature extractor, which support the objective assessment of distances between source and target distributions. Note that if we use the deep feature from the feature extractor of a trained deep UDA network, we cannot objectively assess the distance because its feature extractor has been optimized sufficiently.

We denote by $\tau$ the ratio of transferable parameters among all parameters of dividable modules. $\tau$ is defined by 
\begin{equation}
    \tau = 1 - \left(\frac{\mathrm{1} }{\mathrm{1} + e^{-d_\mathcal{A}}}\right)^2\,, \,\tau > M\,,
    \label{tau}
\end{equation}
where $0<M<1$ is a constant. If the value of proxy $\mathcal{A}$-distance is large, the domain-invariant information becomes less, and the ratio of transferable parameters is small. Note that to avoid the ratio being zero or very close to zero, we should guarantee the ratio is large than $M$. On the contrary, if the value of proxy $\mathcal{A}$-distance is small, the domain-invariant information is sufficient, and the ratio of transferable parameters is large, \eg, one or very close to one for identical distributions.

Accordingly, the number of the transferable parameters of the feature extractor module or the source hypothesis module is defined as follows,
\begin{equation}
    m_f^t = \tau \times m_f, \, \text{and} \,\, m_c^t = \tau \times m_c\,.
    \label{m_f}
\end{equation}
For the domain discrepancy measurement module, there are two cases. If the domain discrepancy measurement module is in the non-adversarial way or the adversarial way, the number of the transferable parameters is respectively defined by
\begin{equation}
    m_d^t = \tau \times m_d, \, \text{or} \,\, m_d^t = (1-\tau) \times m_d\,.
    \label{m_d}
\end{equation}

In each iteration, for each parameter $w_i, i \in [m_\ast]$, the transferable parameters are determined by the result of numerical sorting of $\mathcal{T}_i$, which has been explained before. We denote by the transferable and untransferable parameters of a module by $\mathbf{\mathcal{W}}_\ast^{\text{tr}}$ and $\mathbf{\mathcal{W}}_\ast^{\text{utr}}$ respectively.

\begin{algorithm}[ht]
    \caption{Transferable Parameter Learning.}\label{algorithm}
    \SetKwData{Shuffle}{shuffle}\SetKwData{This}{this}\SetKwData{Up}{up}
    \SetKwFunction{Union}{Union}\SetKwFunction{FindCompress}{FindCompress}
    \SetKwInOut{Input}{input}\SetKwInOut{Output}{output}
    \Input{parameters $\mathcal{W}_f$, $\mathcal{W}_d$, $\mathcal{W}_c$, training sets $\hat{p}$ and $\hat{q}$, learning rate $\eta$, weight decay coefficient $\lambda$, fixed $M$, max epoch $E$, max iteration $T$, max epoch of distribution distance calculation $E'$, $E'\ll E$}
    \Output{parameters $\mathcal{W}_f$, $\mathcal{W}_d$, $\mathcal{W}_c$ after update}
    /* stage 1: determine the ratio of transferable parameters */ \\
    \textbf{initialize} parameters $\mathcal{W}_f$ from ImageNet pretrained model, and parameters $\mathcal{W}_d$ randomly\\
    \For{e' = 1, 2, $\dots$, $E'$}{
        \textbf{optimize} the parameters $\mathcal{W}_d$ of the domain discriminator $D$ \\ 
    }
    \textbf{calculate} the proxy $d_\mathcal{A}$ distance with Eq.~\eqref{d_A} \\
    \textbf{calculate} the ratio $\tau$ with Eq.~\eqref{tau} \\
    /* stage 2: train the deep UDA network with transferable parameters */ \\
    \textbf{initialize} parameters $\mathcal{W}_f$, $\mathcal{W}_d$, $\mathcal{W}_c$ \\
    \For{e = 1, 2, $\dots$, $E$}{
        \textbf{shuffle} training sets $\hat{p}$ and $\hat{q}$\\
    \For{t = 1, 2, $\dots$, $T$}{
        \textbf{fetch} batch $\bar{p}$ from $\hat{p}$, $\bar{q}$ from $\hat{q}$ \\
        /*identify transferable parameters*/\\
        \textbf{divide} $\mathbf{\mathcal{W}}_f$ into $\mathbf{\mathcal{W}}_f^{\text{tr}}$ and $\mathbf{\mathcal{W}}_f^{\text{utr}}$ with Eq.~\eqref{criterion} and Eq.~\eqref{m_f}\\
        \textbf{divide} $\mathbf{\mathcal{W}}_c$ into $\mathbf{\mathcal{W}}_c^{\text{tr}}$ and $\mathbf{\mathcal{W}}_c^{\text{utr}}$ with Eq.~\eqref{criterion} and Eq.~\eqref{m_f}\\
        \textbf{divide} $\mathbf{\mathcal{W}}_d$ into $\mathbf{\mathcal{W}}_d^{\text{tr}}$ and $\mathbf{\mathcal{W}}_d^{\text{utr}}$ with Eq.~\eqref{criterion} and Eq.~\eqref{m_d}\\
        /*update transferable parameter using the robust positive update*/ \\
        \textbf{update} $\mathbf{\mathcal{W}}_f^{\text{tr}}$, $\mathbf{\mathcal{W}}_c^{\text{tr}}$, and $\mathbf{\mathcal{W}}_d^{\text{tr}}$ with Eq.~\eqref{pos}\\
        /*update untransferable parameter using the negative update*/ \\
        \textbf{update} $\mathbf{\mathcal{W}}_f^{\text{utr}}$, $\mathbf{\mathcal{W}}_c^{\text{utr}}$, and $\mathbf{\mathcal{W}}_d^{\text{utr}}$ with Eq.~\eqref{neg}\\
    }}
\end{algorithm}

\subsection{Updating Parameters with Different Update Rules}

We use different update rules for these two types of parameters of the three modules simultaneously inspired by~\cite{DBLP:conf/iclr/XiaL00WGC21}. On the one hand, we perform the robust positive update using the gradients derived from the objective function and weight decay for the transferable parameters:
\begin{equation}
    \mathbf{\mathcal{W}}_\ast^{\text{tr}} (t+1) \rightarrow \mathbf{\mathcal{W}}_\ast^{\text{tr}}(t) - \eta\left(\frac{\partial (\mathcal{L}(\mathbf{\mathcal{W}}_\ast^{\text{tr}}(t)))}{\partial \mathbf{\mathbf{\mathcal{W}}}_\ast^{\text{tr}}(t)} - \lambda \text{sgn}(\mathbf{\mathcal{W}}_\ast^{\text{tr}}(t))\right)\,.
    \label{pos}
\end{equation}

On the other hand, we perform the negative update for the untransferable parameters. Their values are penalized with the weight decay and without the gradients derived from the corresponding objective functions. The negative update is defined by

\begin{equation}
    \mathbf{\mathcal{W}}_\ast^{\text{utr}} (t+1) \rightarrow \mathbf{\mathcal{W}}_\ast^{\text{utr}}(t) - \eta\left(\lambda \text{sgn}(\mathbf{\mathcal{W}}_\ast^{\text{utr}}(t))\right)\,.
    \label{neg}
\end{equation}


The robust positive update applies the gradients derived from the deep UDA objective functions to update the transferable parameters, which can help deep UDA networks learn domain-invariant information. The untransferable parameters tend to over-adapt to domain-specific information, and their gradients are misleading to the generalization of the deep UDA networks. Therefore, we only use weight decay to plenary their values to zero. As they are deactivated, they will not offer to the learning of domain-specific information. The advantage of these two update rules allows us to reduce the side effects of domain-specific information and enhance the learning of domain-invariant information. Algorithm~\ref{algorithm} summarizes the overall procedure of TransPar for optimizing different parameters with separate update rules. Note that TransPar does not change the structure of the deep UDA networks, so the time complexity and space complexity of our algorithm are the same as the original networks. Calculating the ratio between transferable and untransferable parameters takes a little time, but the result is reusable. The different update rules for the two parameters are achieved based on SGD. Thus they spent the same time as the original SGD algorithm.

\begin{table*}[ht]
    \centering
    \caption{Accuracy (\%) on \textbf{Office-31} for unsupervised domain adaptation (ResNet-50).}
    \scalebox{1.0}{
    \begin{tabular}{l| cccccc|l}
    \toprule
        Method & A$\rightarrow$W & D$\rightarrow$W & W$\rightarrow$D & A$\rightarrow$D & D$\rightarrow$A & W$\rightarrow$A & Avg \\ \midrule
        ResNet-50~\cite{DBLP:conf/cvpr/HeZRS16} & 68.4$\pm$0.2 & 96.7$\pm$0.1 & 99.3$\pm$0.1 & 68.9$\pm$0.2 & 62.5$\pm$0.3 & 60.7$\pm$0.3 & 76.1 \\ 
        \textbf{TransPar-ResNet-50} & 77.0$\pm$0.2 & 96.1$\pm$0.1 & 99.0$\pm$0.3 & 81.7$\pm$0.2 & 65.1$\pm$0.1 & 64.4$\pm$0.2 & 80.6 $\uparrow$\\ \midrule

        DAN~\cite{DBLP:conf/icml/LongC0J15} & 80.5$\pm$0.4 & 97.1$\pm$0.2 & 99.6$\pm$0.1 & 78.6$\pm$0.2 & 63.6$\pm$0.3 & 62.8$\pm$0.2 & 80.4 \\ 
        \textbf{TransPar-DAN} & 87.9$\pm$0.5 & 98.7$\pm$0.1 & 100$\pm$0.0 & 86.9$\pm$0.2 & 67.6$\pm$0.3 & 64.5$\pm$0.2 & 84.3 $\uparrow$\\ \midrule

        DANN~\cite{DBLP:conf/icml/GaninL15} & 82.0$\pm$0.4 & 96.9$\pm$0.2 & 99.1$\pm$0.1 & 79.7$\pm$0.4 & 68.2$\pm$0.4 & 67.4$\pm$0.5 & 82.2 \\
        \textbf{TransPar-DANN} & 92.3$\pm$0.5 & 98.9$\pm$0.1  & 100$\pm$0.0  & 89.8$\pm$0.2  & 74.9$\pm$0.3  & 72.3$\pm$0.4  & 88.0 $\uparrow$\\ \midrule


        JAN~\cite{DBLP:conf/nips/LongZ0J16} & 85.4$\pm$0.3 & 97.4$\pm$0.2 & 99.8$\pm$0.2 & 84.7$\pm$0.3 & 68.6$\pm$0.3 & 70.0$\pm$0.4 & 84.3 \\
        \textbf{TransPar-JAN} & 93.2$\pm$0.4 & 98.0$\pm$0.2 & 100$\pm$0.0 & 89.6$\pm$0.3 & 70.7$\pm$0.2 & 71.0$\pm$0.1 & 87.1 $\uparrow$\\ \midrule


        MCD~\cite{DBLP:conf/cvpr/SaitoWUH18} & 88.6$\pm$0.2 & 98.5$\pm$0.1 & 100$\pm$0.0 & 92.2$\pm$0.2 & 69.5$\pm$0.1 & 69.7$\pm$0.3 & 86.5 \\
        \textbf{TransPar-MCD} & 90.1$\pm$0.2 & 98.4$\pm$0.1 & 100$\pm$0.0 & 89.2$\pm$0.3 & 71.6$\pm$0.2 & 70.8$\pm$0.1 & 86.7 $\uparrow$ \\ \midrule

        CDAN~\cite{DBLP:conf/nips/LongC0J18} & 94.1$\pm$0.1 & 98.6$\pm$0.1 & 100$\pm$0.0 & 92.9$\pm$0.2 & 71.0$\pm$0.3 & 69.3$\pm$0.3 & 87.7 \\
        \textbf{TransPar-CDAN} & 93.8$\pm$0.1 & 98.6$\pm$0.1 & 100$\pm$0.0 & 93.3$\pm$0.2 & 72.8$\pm$0.1 & 71.5$\pm$0.2 & 88.3 $\uparrow$\\ \midrule

        MDD~\cite{DBLP:conf/icml/0002LLJ19} & 94.5$\pm$0.3 & 98.4$\pm$0.1 & 100$\pm$0.0 & 93.5$\pm$0.2 & 74.6$\pm$0.3 & 72.2$\pm$0.1 & 88.9 \\ 
        \textbf{TransPar-MDD} & \textbf{95.5$\pm$0.2}  & \textbf{98.9$\pm$0.2}  & \textbf{100$\pm$0.0}  & \textbf{94.2$\pm$0.1}  & \textbf{77.7$\pm$0.2}  & \textbf{72.8$\pm$0.1}  & \textbf{89.9 $\uparrow$}\\ \bottomrule
    \end{tabular}
    \label{tab:office-31}}
\end{table*}

\begin{table*}[ht]
    \centering
    \caption{Accuracy (\%) on \textbf{Office-Home} for unsupervised domain adaptation (ResNet-50).}
    \scalebox{0.9}{
    \begin{tabular}{l|cccccccccccc|l}
    \toprule
        Method & Ar$\rightarrow$Cl & Ar$\rightarrow$Pr & Ar$\rightarrow$Rw & Cl$\rightarrow$Ar & Cl$\rightarrow$Pr & Cl$\rightarrow$Rw & Pr$\rightarrow$Ar & Pr$\rightarrow$Cl & Pr$\rightarrow$Rw & Rw$\rightarrow$Ar & Rw$\rightarrow$Cl & Rw$\rightarrow$Pr & Avg \\ \midrule
        ResNet-50~\cite{DBLP:conf/cvpr/HeZRS16}  & 34.9 & 50.0 & 58.0 & 37.4 & 41.9 & 46.2 & 38.5 & 31.2 & 60.4 & 53.9 & 41.2 & 59.9 & 46.1 \\
        DAN~\cite{DBLP:conf/icml/LongC0J15} & 43.6 & 57.0 & 67.9 & 45.8 & 56.5 & 60.4 & 44 & 43.6 & 67.7 & 63.1 & 51.5 & 74.3 & 56.3 \\
        DANN~\cite{DBLP:conf/icml/GaninL15} & 45.6 & 59.3 & 70.1 & 47.0 & 58.5 & 60.9 & 46.1 & 43.7 & 68.5 & 63.2 & 51.8 & 76.8 & 57.6 \\
        JAN~\cite{DBLP:conf/nips/LongZ0J16} & 45.9 & 61.2 & 68.9 & 50.4 & 59.7 & 61.0 & 45.8 & 43.4 & 70.3 & 63.9 & 52.4 & 76.8 & 58.3 \\
        CDAN~\cite{DBLP:conf/nips/LongC0J18} & 50.7 & 70.6 & 76.0 & 57.6 & 70.0 & 70.0 & 57.4 & 50.9 & 77.3 & 70.9 & 56.7 & 81.6 & 65.8 \\
        MDD~\cite{DBLP:conf/icml/0002LLJ19} & 54.9 & 73.7 & 77.8 & 60.0 & 71.4 & 71.8 & 61.2 & 53.6 & 78.1 & 72.5 & 60.2 & 82.3 & 68.1 \\ \midrule
        \textbf{TransPar-DANN} & 55.1 & 70.4 & 77.8 & 61.6 & 70.8 & 72.1 & 62.1 & \textbf{55.1} & \textbf{80.8} & 73.8 & 60.0 & 83.0 & 68.6 $\uparrow$ \\
        \textbf{TransPar-MDD} & \textbf{55.3} & \textbf{75.9} & \textbf{79.2} & \textbf{63.2} & \textbf{72.4} & \textbf{73.8} & \textbf{63.2} & \textbf{55.1} & 80.0 & \textbf{74.5} & \textbf{60.9} & \textbf{83.7} & \textbf{69.8} $\uparrow$ \\ \bottomrule
    \end{tabular}}
    \label{tab:office-home}
\end{table*}

\section{Experiments}
\label{Experiment}

To verify the feasibility of the proposed learning method in UDA, we thoroughly evaluate the performance on some image classification and regression tasks (keypoint detection) involving six benchmark datasets against state-of-the-art deep UDA methods. The code is available at \url{https://github.com/zhyhan/TransPar}.

\subsection{Experiment on Image Classification}


\subsubsection{Setup}

\begin{figure}[h!]
    \centering
    \includegraphics[width=0.5\textwidth]{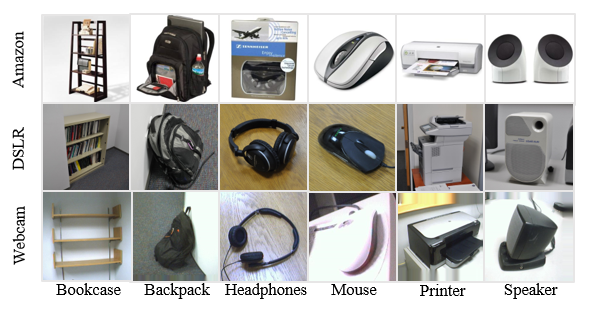}
    \caption{Some example images in the Office-31 dataset.}
    \label{Office-31}
  \end{figure}

\textbf{Office-31}~\cite{DBLP:conf/eccv/SaenkoKFD10} is a standard UDA dataset consisting of three distinct domains, \textbf{A}mazon from the Amazon website, \textbf{W}ebcam by the web camera, and \textbf{D}SLR by digital SLR camera (see Fig.~\ref{Office-31}). It has 4,652 images with 31 unbalanced classes.

\begin{figure}[h!]
    \centering
    \includegraphics[width=0.48\textwidth]{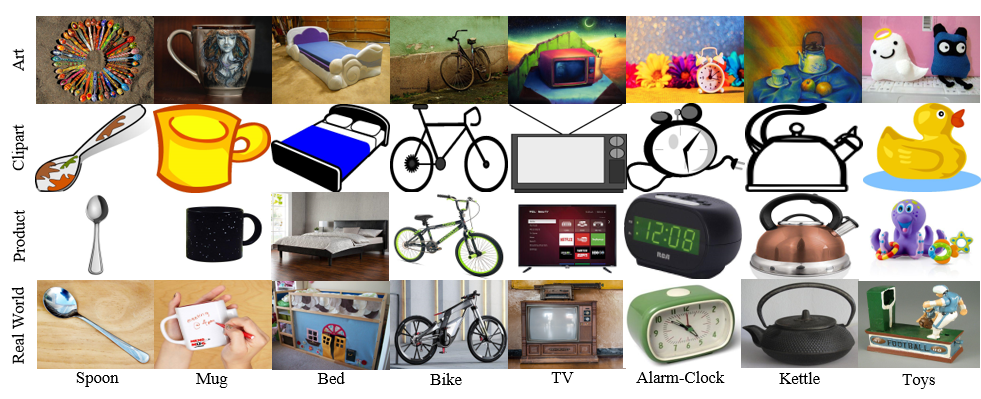}
    \caption{Some example images in the Office-Home dataset.}
    \label{Office-Home}
  \end{figure}

\textbf{Office-Home}~\cite{DBLP:conf/cvpr/VenkateswaraECP17} is a more challenging domain adaptation dataset consisting of 15,599 images with 65 unbalanced classes. It consists of four more distinct domains: \textbf{Ar}tistic images, \textbf{Cl}ip Art, \textbf{Pr}oduct images, and \textbf{R}eal-\textbf{w}orld images (see Fig.~\ref{Office-Home}).

\begin{figure}[h!]
    \centering
    \includegraphics[width=0.5\textwidth]{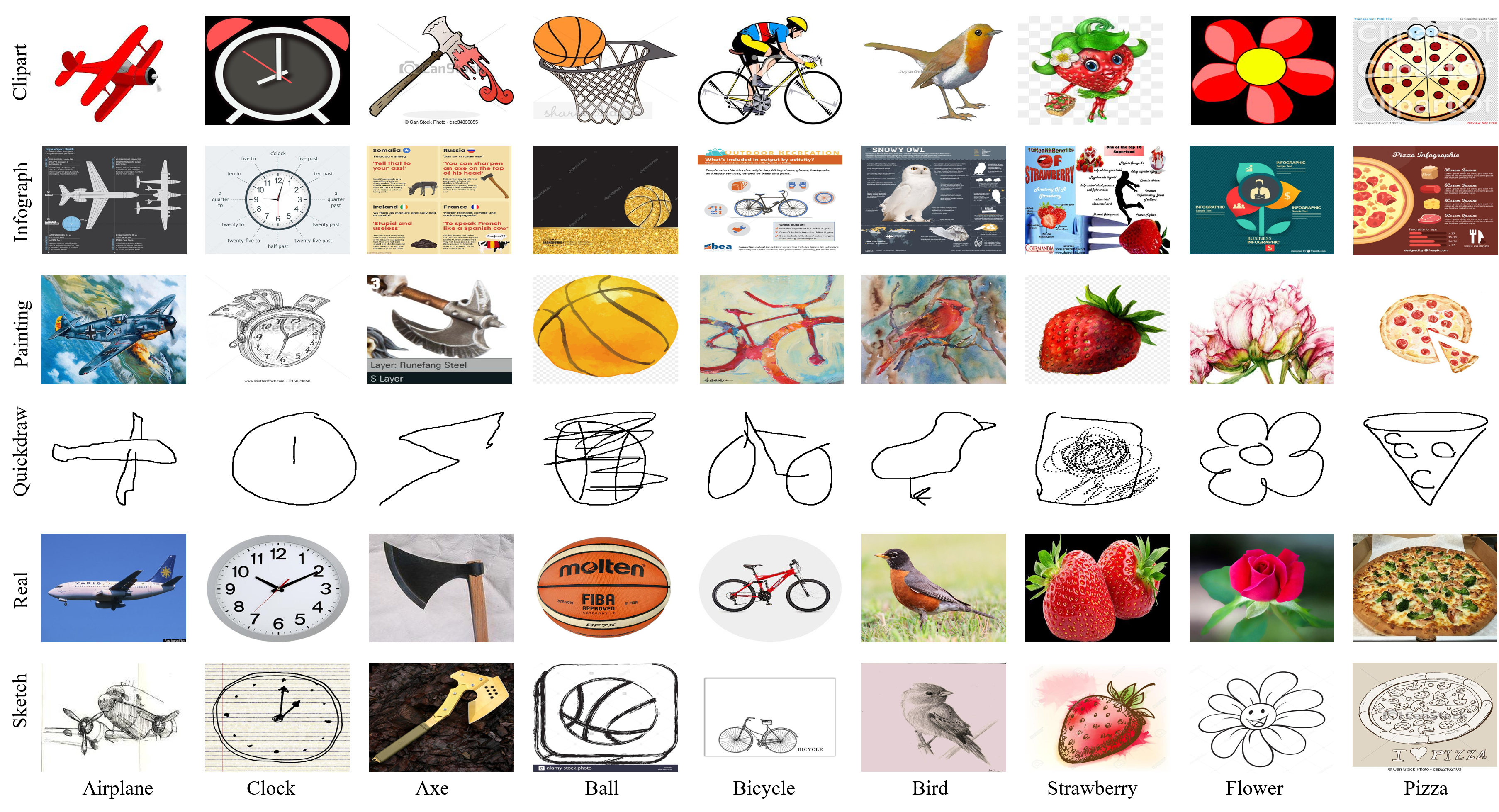}
    \caption{Some example images in the DomainNet dataset.}
    \label{DomainNet}
  \end{figure}

\textbf{DomainNet}~\cite{DBLP:conf/iccv/PengBXHSW19} is the most challenging domain adaptation dataset consisting of six different domains: \textbf{c}lipart collected from clipart images, \textbf{r}eal collected from photo-realistic or real-world images, \textbf{s}ketch collected from the sketches of specific objects, \textbf{i}nfographic images with specific object, \textbf{p}ainting artistic depictions of objects in the form of paintings and quickdraw, and \textbf{q}uickdraw collected from the drawings of game (see Fig.~\ref{DomainNet}). This dataset contains about 600,000 images distributed in 345 categories.

\begin{figure}[h!]
    \centering
    \includegraphics[width=0.5\textwidth]{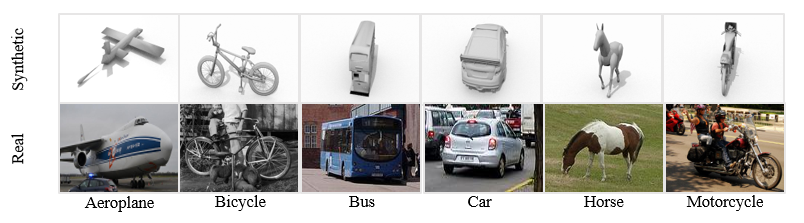}
    \caption{Some example images in the VisDA-2017 dataset.}
    \label{VisDA}
\end{figure}

\textbf{VisDA-2017}~\cite{visda2017} is a simulation-to-real dataset with two extremely distinct domains: \textbf{Synthetic} 2D renderings of 3D models generated from different angles and with different lighting conditions, and \textbf{Real} collected from photo-realistic or real-world image datasets (see Fig.~\ref{VisDA}). With 280,000 images in 12 classes, the scale of VisDA-2017 brings challenges to domain adaptation.

\begin{table*}[h!]
    \centering
    \caption{Accuracy (\%) on \textbf{DomainNet} for unsupervised domain adaptation (ResNet-101). In each sub-table, the column-wise domains are selected as the source domain, and the row-wise domains are selected as the target domain.}
    \scalebox{0.6}{
    \begin{tabular}{l|lllllll||l|lllllll||l|lllllll||l|lllllll}
      \toprule
      \textbf{AlexNet} & clp & inf & pnt & qdr & rel & skt & Avg & \textbf{DAN} & clp & inf & pnt & qdr & rel & skt & Avg & \textbf{JAN} & clp & inf & pnt & qdr & rel & skt & Avg & \textbf{DANN} & clp & inf & pnt & qdr & rel & skt & Avg \\ \midrule
      clp & 65.5  & 8.2  & 21.4  & 10.5  & 36.1  & 10.8  & 17.4  & clp & N/A & 9.1  & 23.4  & 16.2  & 37.9  & 29.7  & 23.2  & clp & N/A & 7.8  & 24.5  & 14.3  & 38.1  & 25.7  & 22.1  & clp & N/A & 9.1  & 23.2  & 13.7  & 37.6  & 28.6  & 22.4  \\
      inf & 32.9  & 27.7  & 23.8  & 2.2  & 26.4  & 13.7  & 19.8  & inf & 17.2  & N/A & 15.6  & 4.4  & 24.8  & 13.5  & 15.1  & inf & 17.6  & N/A & 18.7  & 8.7  & 28.1  & 15.3  & 17.7  & inf & 17.9  & N/A & 16.4  & 2.1  & 27.8  & 13.3  & 15.5  \\
      pnt & 28.1  & 7.5  & 57.6  & 2.6  & 41.6  & 20.8  & 20.1  & pnt & 29.9  & 8.9  & N/A & 7.9  & 42.1  & 26.1  & 23.0  & pnt & 27.5  & 8.2  & N/A & 7.1  & 43.1  & 23.9  & 22.0  & pnt & 29.1  & 8.6  & N/A & 5.1  & 41.5  & 24.7  & 21.8  \\
      qdr & 13.4  & 1.2  & 2.5  & 68.0  & 5.5  & 7.1  & 5.9  & qdr & 14.2  & 1.6  & 4.4  & N/A & 8.5  & 10.1  & 7.8  & qdr & 17.8  & 2.2  & 7.4  & N/A & 8.1  & 10.9  & 9.3  & qdr & 16.8  & 1.8  & 4.8  & N/A & 9.3  & 10.2  & 8.6  \\
      rel & 36.9  & 10.2  & 33.9  & 4.9  & 72.8  & 23.1  & 21.8  & rel & 37.4  & 11.5  & 33.3  & 10.1  & N/A & 26.4  & 23.7  & rel & 33.5  & 9.1  & 32.5  & 7.5  & N/A & 21.9  & 20.9  & rel & 36.5  & 11.4  & 33.9  & 5.9  & N/A & 24.5  & 22.4  \\
      skt & 35.5  & 7.1  & 21.9  & 11.8  & 30.8  & 56.3  & 21.4  & skt & 39.1  & 8.8  & 28.2  & 13.9  & 36.2  & N/A & 25.2  & skt & 35.3  & 8.2  & 27.7  & 13.3  & 36.8  & N/A & 24.3  & skt & 37.9  & 8.2  & 26.3  & 12.2  & 35.3  & N/A & 24.0  \\
      Avg & 29.4  & 6.8  & 20.7  & 6.4  & 28.1  & 15.1  & \textbf{17.8}  & Avg & 27.6  & 8.0  & 21.0  & 10.5  & 29.9  & 21.2  & \textbf{19.7}  & Avg & 26.3  & 7.1  & 22.2  & 10.2  & 30.8  & 19.5  & \textbf{19.4}  & Avg & 27.6  & 7.8  & 20.9  & 7.8  & 30.3  & 20.3  & \textbf{19.1}  \\ \midrule \midrule
      \textbf{RTN} & clp & inf & pnt & qdr & rel & skt & Avg & \textbf{ADDA} & clp & inf & pnt & qdr & rel & skt & Avg & \textbf{MCD} & clp & inf & pnt & qdr & rel & skt & Avg & \textbf{TransPar} & clp & inf & pnt & qdr & rel & skt & Avg \\ \midrule
      clp & N/A & 8.1  & 21.1  & 13.1  & 36.1  & 26.5  & 21.0  & clp & N/A & 11.2  & 24.1  & 3.2  & 41.9  & 30.7  & 22.2  & clp & N/A & 14.2  & 26.1  & 1.6  & 45.0  & 33.8  & 24.1  & clp & N/A & 26.2  & 41.7  & 24.8  & 53.1  & 57.6  & 40.7  \\
      inf & 15.6  & N/A & 15.3  & 3.4  & 25.1  & 12.8  & 14.4  & inf & 19.1  & N/A & 16.4  & 3.2  & 26.9  & 14.6  & 16.0  & inf & 23.6  & N/A & 21.2  & 1.5  & 36.7  & 18.0  & 20.2  & inf & 20.5  & N/A & 19.4  & 4.7  & 22.7  & 20.5  & 17.6  \\
      pnt & 26.8  & 8.1  & N/A & 5.2  & 40.6  & 22.6  & 20.7  & pnt & 31.2  & 9.5  & N/A & 8.4  & 39.1  & 25.4  & 22.7  & pnt & 34.4  & 14.8  & N/A & 1.9  & 50.5  & 28.4  & 26.0  & pnt & 38.5  & 27.7  & N/A & 10.2  & 51.8  & 47.0  & 35.0  \\
      qdr & 15.1  & 1.8  & 4.5  & N/A & 8.5  & 8.9  & 7.8  & qdr & 15.7  & 2.6  & 5.4  & N/A & 9.9  & 11.9  & 9.1  & qdr & 15.0  & 3.0  & 7.0  & N/A & 11.5  & 10.2  & 9.3  & qdr & 13.3  & 4.7  & 0.0  & N/A & 10.1  & 12.0  & 8.0  \\
      rel & 35.3  & 10.7  & 31.7  & 7.5  & N/A & 22.9  & 21.6  & rel & 39.5  & 14.5  & 29.1  & 12.1  & N/A & 25.7  & 24.2  & rel & 42.6  & 19.6  & 42.6  & 2.2  & N/A & 29.3  & 27.2  & rel & 56.2  & 44.0  & 55.5  & 17.4  & N/A & 55.8  & 45.8  \\
      skt & 34.1  & 7.4  & 23.3  & 12.6  & 32.1  & N/A & 21.9  & skt & 35.3  & 8.9  & 25.2  & 14.9  & 37.6  & N/A & 25.4  & skt & 41.2  & 13.7  & 27.6  & 3.8  & 34.8  & N/A & 24.2  & skt & 45.2  & 25.9  & 38.1  & 20.4  & 39.4  & N/A & 33.8  \\
      Avg & 25.4  & 7.2  & 19.2  & 8.4  & 28.4  & 18.7  & \textbf{17.9}  & Avg & 28.2  & 9.3  & 20.1  & 8.4  & 31.1  & 21.7  & \textbf{19.8}  & Avg & 31.4  & 13.1  & 24.9  & 2.2  & 35.7  & 23.9  & 21.9  & Avg & 34.7  & 25.7  & 30.9  & 15.5  & 35.4  & 38.6  & \textbf{30.1} \\
      \bottomrule
    \end{tabular}}
    \label{tab:domainnet}
  \end{table*}

We compare our designed \textbf{TransPar} algorithm with state-of-the-art methods: \textbf{ResNet-50}~\cite{DBLP:conf/cvpr/HeZRS16}, Deep Adaptation Network~(\textbf{DAN})~\cite{DBLP:conf/icml/LongC0J15}, Domain Adversarial Network~(\textbf{DANN})~\cite{DBLP:conf/icml/GaninL15}, Adversarial Discriminative Domain Adaptation~(\textbf{ADDA})~\cite{tzeng2017adversarial}, Joint Adaptation Network~(\textbf{RTN})~\cite{DBLP:conf/nips/LongZ0J16}, Generate to Adapt (\textbf{GTA})~\cite{DBLP:conf/cvpr/Sankaranarayanan18a}, Maximum Classifier Discrepancy (\textbf{MCD})~\cite{DBLP:conf/cvpr/SaitoWUH18}, Conditional Domain Adversarial Network (\textbf{CDAN})~\cite{DBLP:conf/nips/LongC0J18}, and Margin Disparity Discrepancy based algorithm~(\textbf{MDD})~\cite{DBLP:conf/icml/0002LLJ19}. Note that TransPar-$\ast$ denotes that we have integrated our TransPar algorithm into an existing algorithm, \eg, TransPar-MDD, TransPar-DANN.

We follow the commonly used experimental protocol of UDA from~\cite{DBLP:conf/icml/GaninL15,DBLP:conf/icml/0002LLJ19}. We implement our algorithm in Pytorch~\cite{DBLP:conf/nips/PaszkeGMLBCKLGA19}. We report the average accuracies of five independent experiments. The base models depend on the specific algorithm or dataset, \eg, ResNet-50 for Office-31 and Office-Home, ResNet-101 for VisDA-2017 and DomainNet. The base models with parameters are pre-trained from ImageNet. The domain discriminator used for measuring the $\mathcal{A}$-distance is a $2$-layer fully-connected neural network. We set the early training epoch $E'$ to $10$. We use 80\% data to train the domain discriminator and 20\% data to calculate the $\mathcal{A}$-distance. We set $M$ to 0.1 to prevents all parameters from becoming inactive. We set the max epoch $E$ to 30.  We use mini-batch SGD with the weight decay coefficient $\lambda$ of 0.75 according to MDD~\cite{DBLP:conf/icml/0002LLJ19}. The initial learning rate is $0.001$. The weight of the entropy loss $\alpha$ is set to $0.1$ according to CDAN~\cite{DBLP:conf/nips/LongC0J18}. Data augmentation strategies include the random resized crop and the random horizontal flip. All the compared models are implemented according to the Transfer Learning Library (Dalib)~\cite{dalib}.

\subsubsection{Results}


\textbf{Results on Office-31.} Table~\ref{tab:office-31} reports the results on Office-31. TransPar achieves state-of-the-art accuracies on all six transfer tasks. After incorporating into existing methods, our algorithm outperforms all the original methods, showing its efficacy and universality. For example, when TransPar is built-in the adversarial training based method (TransPar-DANN), it significantly outperforms DANN by a large margin, over 5\% average accuracy. Note that in some large-to-small tasks (A$\rightarrow$W, A$\rightarrow$D), TransPar-DANN improves over 10\% accuracy. This phenomenon indicates that if the data amount of the source domain is larger than the target domain, the UDA model easily learns more source-specific information. Nevertheless, our algorithm can reduce the side effect of domain-specific information in the learning process and thus enhance the learning of domain-invariant information. When TransPar is built-in the moment matching method (\eg, DAN), TransPar-DAN achieves 84.3\% average accuracy, 3.9\% higher than the original DAN. In the large-to-small task (A$\rightarrow$D), TransPar-MDD obtains 86.9\% accuracy, and DAN gets 78.6\% accuracy that shows an 8.3\% accuracy gap. The above results explicitly verify that TransPar properly contributes to learning more domain-invariant information and successfully bridge the domain discrepancy cross domains.

\textbf{Results on Office-Home.}
Table~\ref{tab:office-home} reports the results on Office-Home. We can observe a similar effect: TransPar makes a remarkable performance boost under all 12 tasks with distribution mismatch. Specifically, when TransPar is built-in DANN, it achieves 68.6\% average accuracy while DANN only achieves 57.6\% average accuracy, showing an almost 11\% gap. These results indeed verify the necessity and effectiveness of transferable parameter learning.

\textbf{Results on DomainNet.}
DomainNet supports many real-world domain adaptation tasks involving multiple domains, massive training data, and numerous categories. Therefore, DomainNet can thoroughly verify the capability of our algorithm in practical applications. Table~\ref{tab:domainnet} summarizes the results on DomainNet where our algorithm (TransPar-DANN) delivers state-of-the-art accuracies on all the complex UDA tasks. We can observe that TransPar-DANN furthermore obtains performance gain (an improvement of over 10\% accuracy) when the distribution discrepancy is enormous, verifying that the transferable parameter learning is feasible in real-world domain adaptation tasks.

\begin{table}[t]
  \centering
  \caption{Accuracy (\%) on \textbf{VisDA-2017} (ResNet-101).}
  \scalebox{1.0}{
  \begin{tabular}{lc}
  \toprule
      Method & Synthetic$\rightarrow$ Real\\ \midrule
      DANN~\cite{DBLP:conf/icml/GaninL15} & 55.6 \\
      JAN~\cite{DBLP:conf/nips/LongZ0J16} & 61.6  \\
      MCD~\cite{DBLP:conf/cvpr/SaitoWUH18} & 69.2 \\
      GTA~\cite{DBLP:conf/cvpr/Sankaranarayanan18a} & 69.5 \\
      CDAN~\cite{DBLP:conf/nips/LongC0J18} & 70.0 \\
      MDD~\cite{DBLP:conf/icml/0002LLJ19} & 74.6 \\ \midrule
      \textbf{TransPar-DANN} & \textbf{73.4 $\uparrow$} \\
      \textbf{TransPar-MDD} & \textbf{76.3} $\uparrow$ \\ \bottomrule
  \end{tabular}}
  \label{tab:VisDA-2017}
\end{table}

\textbf{Results on VisDA-2017.} Table~\ref{tab:VisDA-2017} reports the results on the large-scale dataset VisDA-2017. The significant challenges of VisDA-2017 lie in the considerable distribution shift between the large-scale synthetic domain and the real domain. Nevertheless, TransPar-DANN achieves remarkable performance gain (almost 17.8\%) from 55.6\% to 73.4\%. Also, TransPar-MDD improves the accuracy of MDD from 74.6\% to 76.3\% after a simple adjustment. The above results demonstrate that TransPar performs in large-scale tasks with considerable distribution shifts. These results also confirm that our method possesses both simplicity and performance strength.

\begin{table}[h!]
    \centering
    \caption{Accuracy (\%) of TransPar-MDD with different scope of transferable parameters}
    \scalebox{0.8}{
    \begin{tabular}{c|cccccc|c}
        \toprule
        Method
        &A$\rightarrow$W & W$\rightarrow$A & A$\rightarrow$D & D$\rightarrow$A & W$\rightarrow$D & D$\rightarrow$W & Avg \\
        \midrule
        TransPar+FE & 92.7 & 98.5 & 100 & 93.6 & 76.0 & 70.7 & 88.6 \\ 
        TransPar+SH & 93.5 & 98.3 & 100 & 93.6 & 76.2 & 71.7 & 88.9 \\ 
        TransPar+DD & 94.0 & 98.5 & 100 & 93.4 & 75.9 & 71.2 & 88.8 \\ 
        TransPar+FE+DD & 94.2 & 98.6 & 100 & 93.8 & 76.5 & 72.1 & 89.2 \\ 
        TransPar+FE+SH & 94.6 & 98.6 & 100 & 94.0 & 76.8 & 72.5 & 89.4 \\
        TransPar+FE+SH-EN & 93.8 & 98.4 & 100 & 93.6 & 76.7 & 72.1 & 89.1 \\ \midrule
        TransPar+FE+SH+DD & \textbf{95.5} & \textbf{98.9} & \textbf{100} & \textbf{94.2} & \textbf{77.7} & \textbf{72.8} & \textbf{89.9} \\
        \bottomrule
    \end{tabular}}
    \label{tab:parameter scope}
\end{table}

\subsubsection{Ablation Analysis}

In this section, we dissect the strengths of our algorithm and report the results of our ablation study.

\textbf{Scope of Transferable Parameters.}
Firstly, we validate that all the parameters of the three standard modules in deep UDA networks are better to be divided into transferable and untransferable parameters if we add an unsupervised objective function to the target data. Table~\ref{tab:parameter scope} reports the results of DANN with different scopes of transferable parameters: TransPar+FE (TransPar on feature extractor only), TransPar+SH (TransPar on source hypothesis only), TransPar+DD (TransPar on domain discriminator classifier only), TransPar+FE+SH (TransPar on feature extractor and source hypothesis), TransPar+FE+DD (TransPar on feature extractor and domain discriminator), TransPar+FE+SH-SE (TransPar on feature extractor and source hypothesis but without entropy loss function), and TransPar+FE+SH+DD (TransPar on all the three modules). These results prove the advantage of TransPar and the necessity of considering all the parameters of the three standard modules in deep UDA networks. Furthermore, we also justify our claim that we need to add unsupervised loss on the source hypothesis. We can observe that adding the entropy loss function achieves a noticeable performance gain. On the one hand, when TransPar performs on any module of deep UDA networks, the average accuracies are improved to a certain extent, which further verifies the universality of TransPar, \ie, it can be applied to any deep UDA networks after a simple adjustment. 

\begin{figure}[t!]
    \centering
    \resizebox{0.5\textwidth}{!}{
        \subfigure[D$\rightarrow$A]{\includegraphics[width=0.25\textwidth]{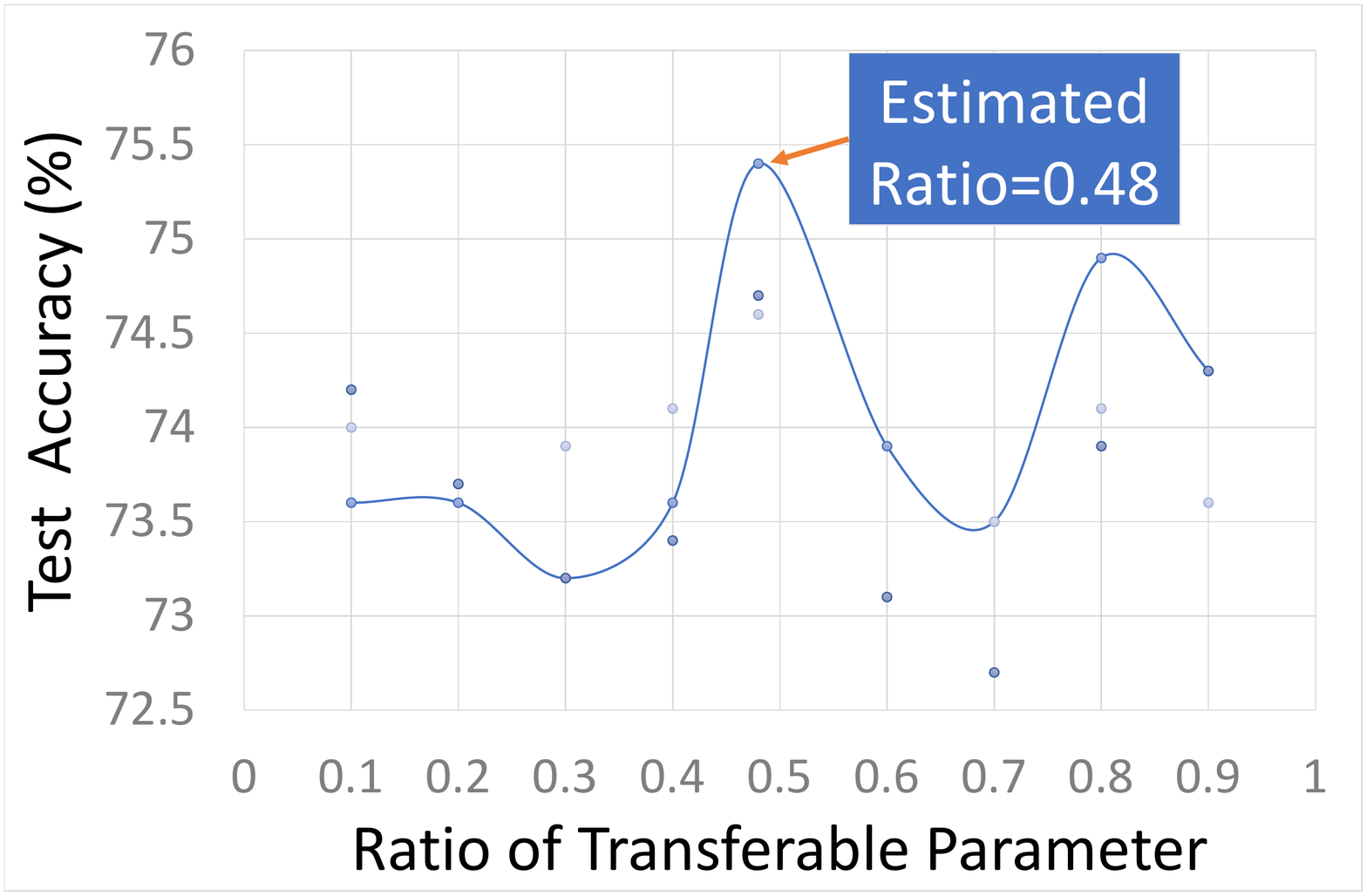}}
        \subfigure[W$\rightarrow$A]{\includegraphics[width=0.25\textwidth]{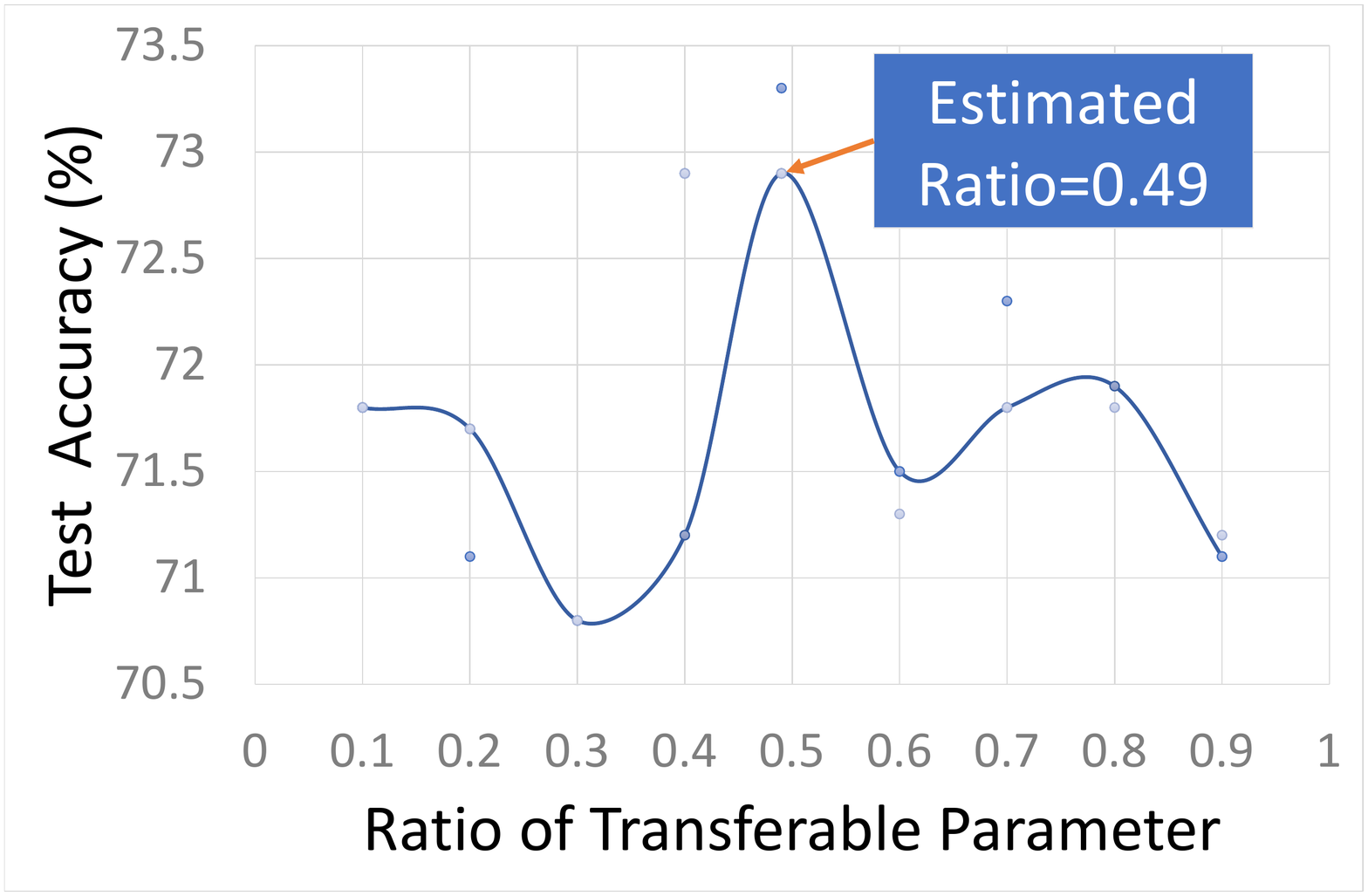}}}
    \caption{Accuracy (\%) of TransPar-DANN based on wider ratios of transferable parameters. The blue dots represent the results of independent experiments.}
    \label{ratio}
\end{figure}

\textbf{Ratio of Transferable Parameters.}
While Table~\ref{tab:office-31} and Table~\ref{tab:office-home} have demonstrated the strengths of our transferable parameters for reducing the side effect of domain-specific information, we provide a broader spectrum for more in-depth analysis for the ratio of transferable parameters. Fig.~\ref{ratio} shows the results of TransPar-MDD on wider ratios of transferable parameters, in which the ratios are set from 0.1 to 0.9 or according to the distribution distance. We can observe that using the estimated ratio maintains the best accuracy by a large margin. Fig.~\ref{ratio} also points that the accuracies produced by different ratios have not large margins, indicating that the performance of transferable parameter learning is not very sensitive to ratios.

\begin{figure}[h!]
    \centering
    \includegraphics[width=0.35\textwidth]{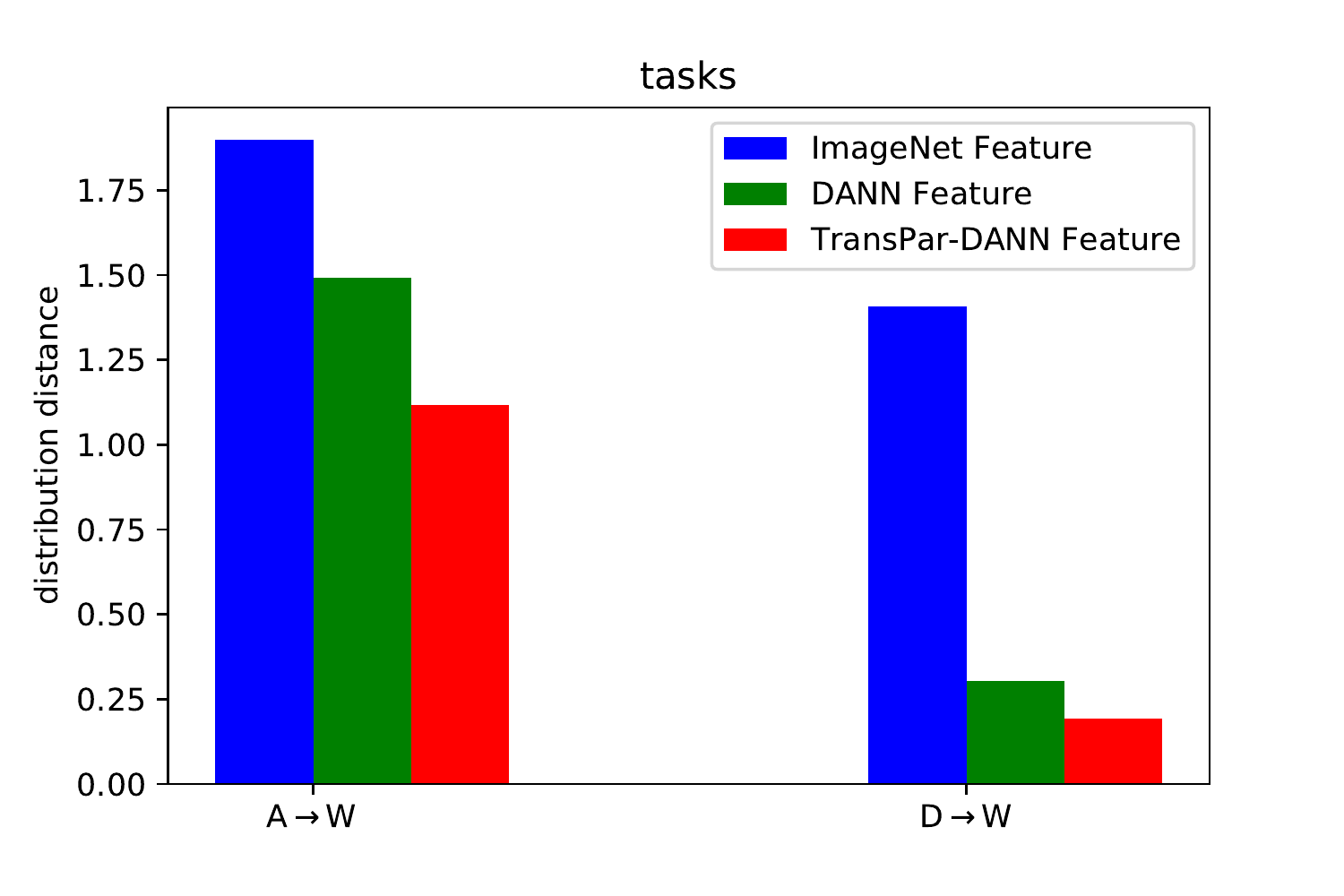}
    \caption{The proxy $\mathcal{A}$-distance calculated with the deep features from ImageNet, DANN, and TransPar-DANN.}
    \label{distance}
\end{figure}

\textbf{Distribution Distance Measurement.}
Fig.~\ref{distance} shows the proxy $\mathcal{A}$-distance values calculated with the deep features from ImageNet, DANN, and TransPar-DANN. We can observe that TransPar-DANN successfully decreases the distribution discrepancy between source and target domains. Fig.~\ref{distance} also shows that with the training of deep UDA networks, the distribution distances cross domains decrease to a certain extent; thus, the trained deep UDA network features cannot be used to measure the actual distribution distance. Therefore, the calculation of distribution distance based on the ImageNet pre-trained features is more suitable than the trained deep UDA network features.

\begin{table}[h!]
    \centering
    \caption{Accuracy (\%) of TransPar-MDD with different update rules.}
    \scalebox{0.75}{
    \begin{tabular}{c|cccccc|c}
        \toprule
        Method
        &A$\rightarrow$W & W$\rightarrow$A & A$\rightarrow$D & D$\rightarrow$A & W$\rightarrow$D & D$\rightarrow$W & Avg \\
        \midrule
        One-shot-start & 93.8 & 98.0 & 100 & 92.4 & 74.5 & 70.9 & 88.3 \\
        One-shot-last & 94.1 & 98.6 & 100 & 93.2 & 74.3 & 71.- & 88.5 \\
        \textbf{Iterative Manner} & \textbf{95.5} & \textbf{98.9} & \textbf{100} & \textbf{94.2} & \textbf{77.7} & \textbf{72.8} & \textbf{89.9} \\
        \midrule
        Weight Decay (weight only) & 94.7 & 98.7 & 100 & 93.4 & 75.8 & 71.5 & 89.0\\
        Weight Decay (gradient only) & 94.2 & 98.7 & 100 & 93.6 & 75.0 & 72.0 & 88.9 \\ 
        \textbf{Weight Decay (both)} & \textbf{95.5} & \textbf{98.9} & \textbf{100} & \textbf{94.2} & \textbf{77.7} & \textbf{72.8} & \textbf{89.9} \\
        \bottomrule
    \end{tabular}}
    \label{tab:update_rule}
\end{table}

\textbf{Differnt Update Rules.}
Table~\ref{tab:update_rule} exhibits the results of different update rules. One-shot-start refers to identifying untransferable parameters at the first iteration (before training) and letting the untransferable parameters' weight value to zero. One-shot-last identifies untransferable parameters at the last iteration (after training) and lets the weight value zero. Weight Decay (weight only) and Weight Decay (gradient only) refer to using the weight value or gradient value to identify the untransferable parameters, respectively. Iterative manner refers to identifying untransferable parameters in each training iteration. Table~\ref{tab:update_rule} shows that the weight decay (both) achieves the best performance compared with various update rules, which coincides with our intuition that it is important to exploit weight and gradient information in the iterative manner during training.

\begin{figure}[h!]
    \centering
    \resizebox{0.5\textwidth}{!}{
        \subfigure[DANN]{\includegraphics[width=0.25\textwidth]{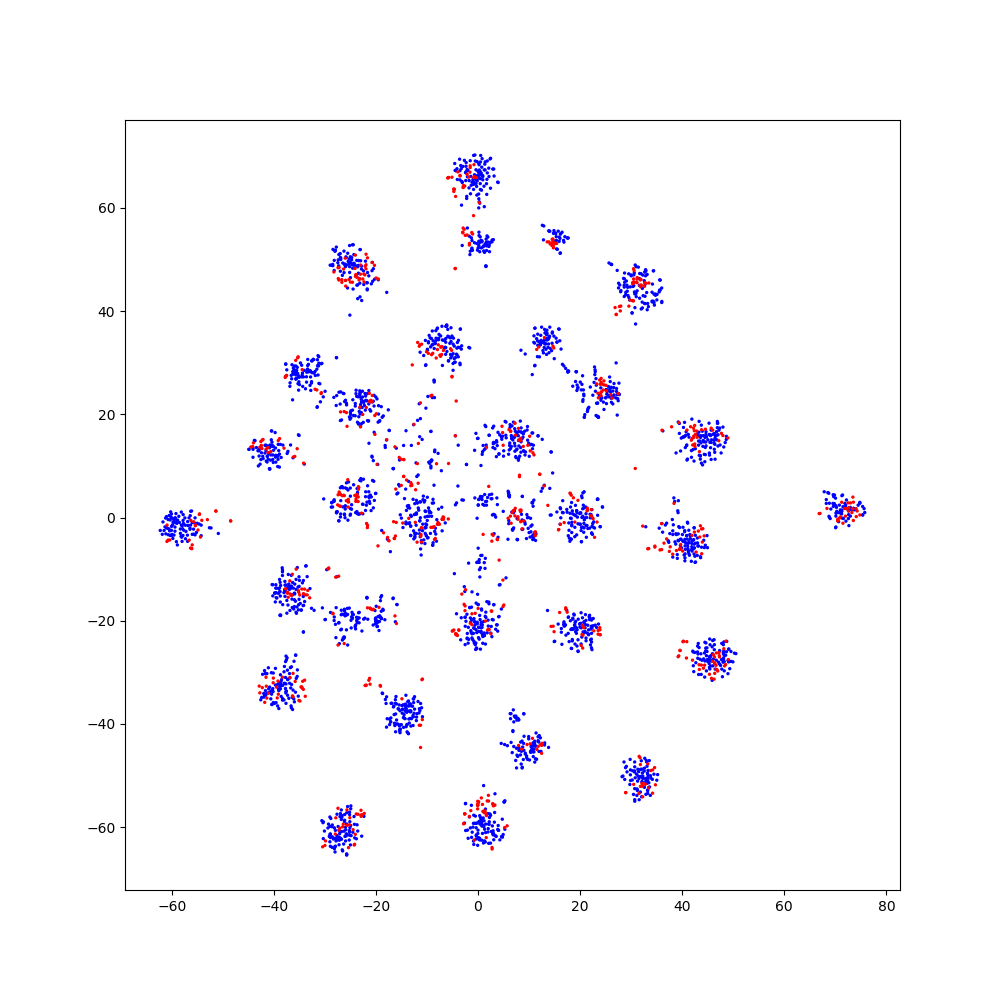}}
        \subfigure[TransPar-DANN]{\includegraphics[width=0.25\textwidth]{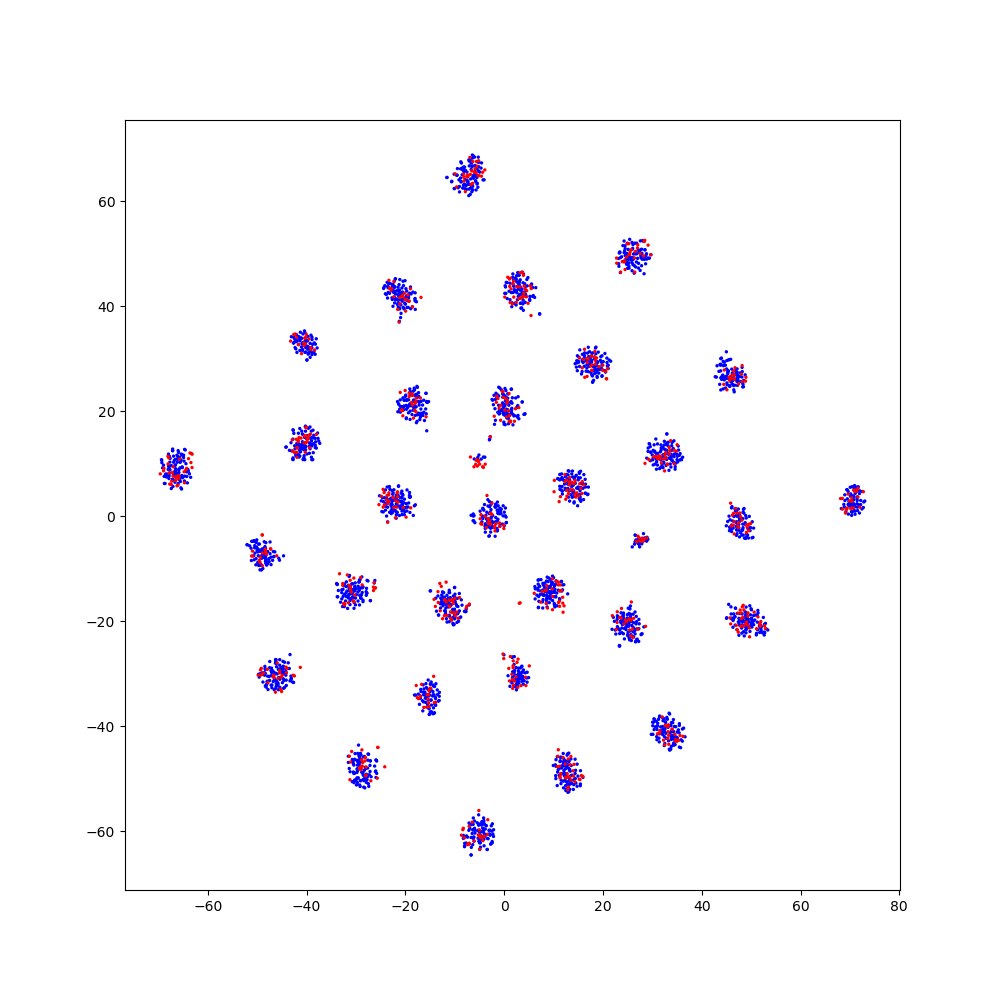}}}
    \caption{The t-SNE visualization of features of examples from each class on the target domain.}
    \label{t-SNE}
\end{figure}

\textbf{Feature Visualization.}
Fig.~\ref{t-SNE} shows the t-SNE embeddings~\cite{DBLP:conf/icml/DonahueJVHZTD14} of the learned features by DANN and TransPar-DANN on the A$\rightarrow$W task. While the features of DANN are mixed up, the features of TransPar-DANN are more discriminative in each class of target examples, which verifies that our method can better learn the domain-invariant representations dominated by the transferable parameters.

\subsection{Experiment on Keypoint Detection}

\subsubsection{Setup}

\begin{figure}[ht]
  \centering
  \includegraphics[width=0.48\textwidth]{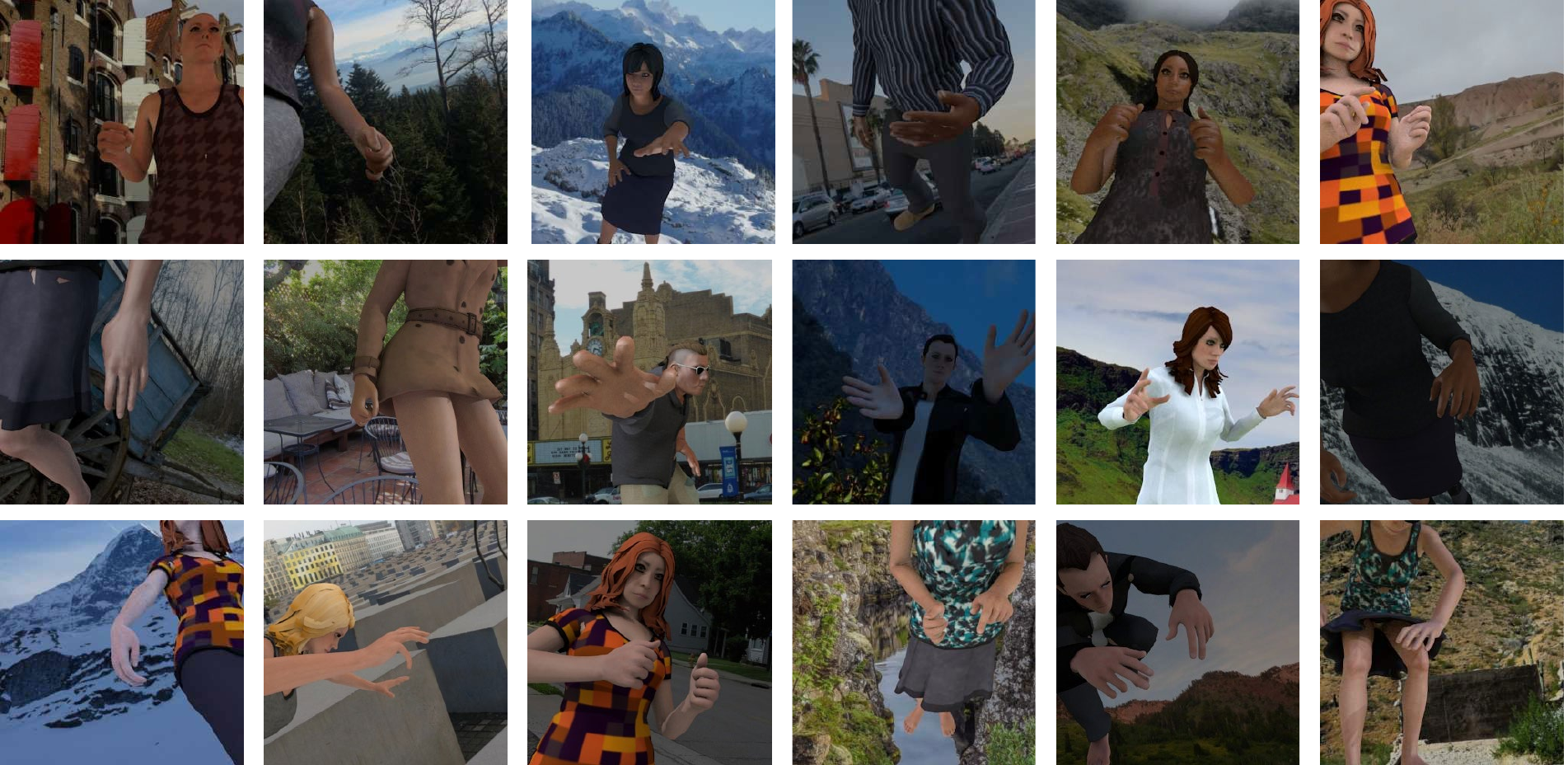}
  \caption{Some example images in the \textit{RHD} dataset.}
  \label{RHD}
\end{figure}

\textbf{Rendered Hand Pose (RHD)}~\cite{zb2017hand} is a synthetic hand keypoint detection dataset, which contains 41,258 synthetic training images and 2,728 synthetic testing images. RHD provides precise annotations for 21 hand keypoints and covers a variety of viewpoints and challenging hand poses. However, hands in this dataset have very different appearances from those in reality (see Fig.~\ref{RHD}).

\begin{figure}[ht]
  \centering
  \includegraphics[width=0.48\textwidth]{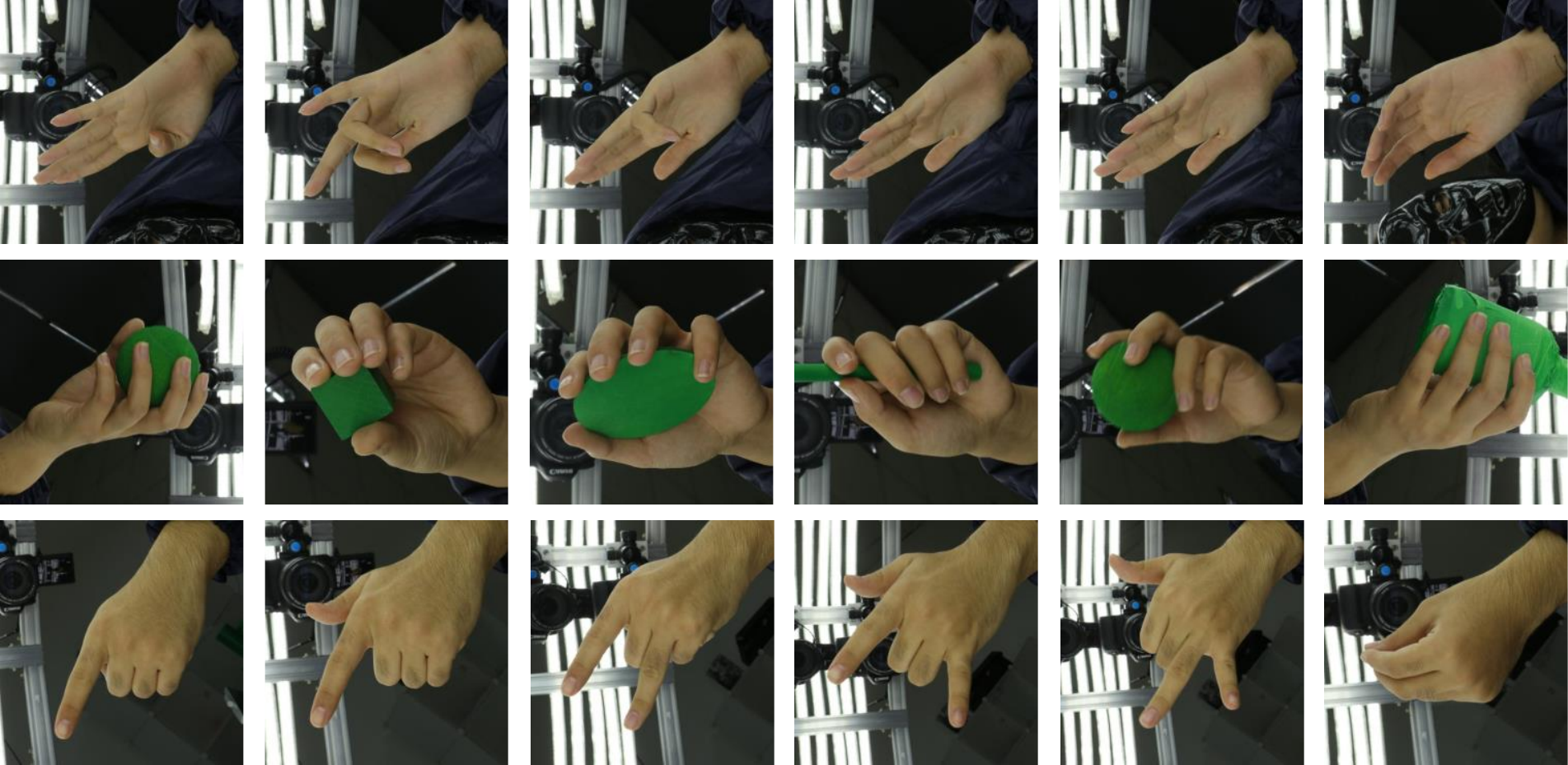}
  \caption{Some example images in the \textit{H3D} dataset.}
  \label{H3D-examples}
\end{figure}

\textbf{Hand-3D-Studio (H3D)}~\cite{DBLP:conf/icassp/ZhaoWXW20} is a real-world hand keypoint detection dataset, which contains 22,000 frames sampled from videos (see Fig.\ref{H3D-examples}). Following~\cite{DBLP:journals/corr/abs-2103-06175}, 3,200 frames are randomly picked as the testing set, and the remaining frames is used as the training set.

\begin{figure*}[ht]
    \centering
    \includegraphics[width=1\textwidth]{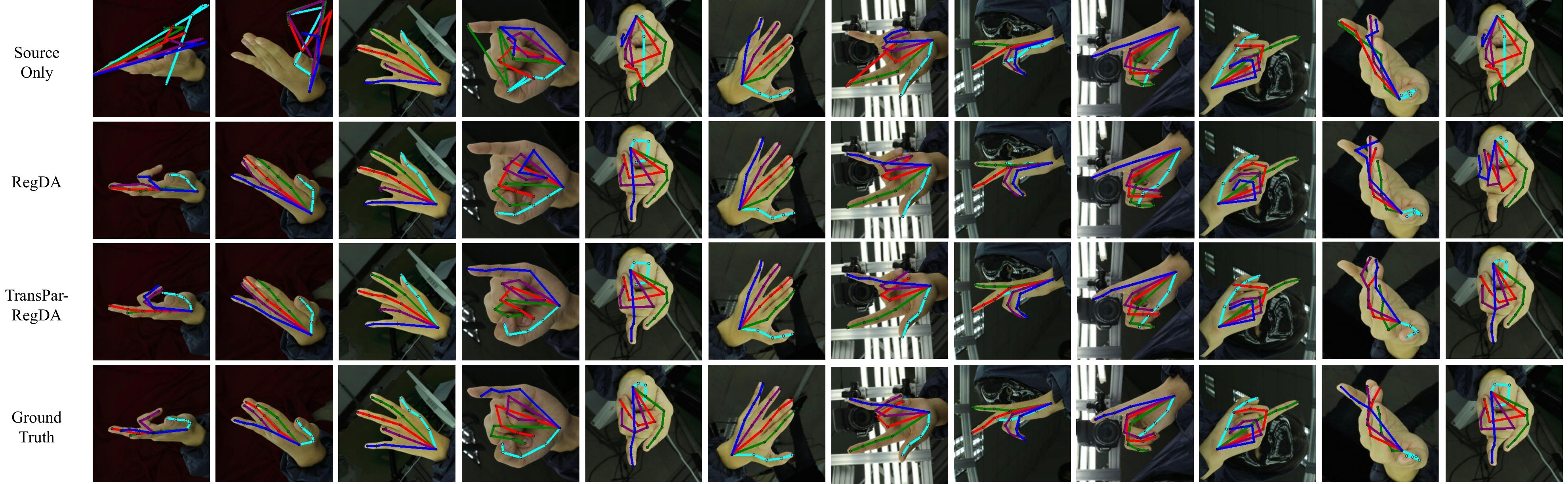}
    \caption{Qualitative results of some images in the \textit{H3D} dataset.}
    \label{H3D}
\end{figure*}

We compare our designed \textbf{TransPar} algorithm with state-of-the-art methods: Source Only (\textbf{ResNet-101}~\cite{DBLP:conf/cvpr/HeZRS16}), Deep Adaptation Network (\textbf{DAN})~\cite{DBLP:conf/icml/LongC0J15}, Domain Adversarial Network (\textbf{DANN})~\cite{DBLP:conf/icml/GaninL15}), Maximum Classifier Discrepancy (\textbf{MCD})~\cite{DBLP:conf/cvpr/SaitoWUH18}, Margin Disparity Discrepancy based algorithm (\textbf{MDD})~\cite{DBLP:conf/icml/0002LLJ19}, and Regressive Domain Adaptation (\textbf{RegDA})~\cite{DBLP:journals/corr/abs-2103-06175}. Note that RegDA is the state-of-the-art method for unsupervised keypoint detection. All methods are pretrained on the source domain for 70 epochs and then finetuned with the unlabeled data on the target domain for 30 epochs. We report the final Percentage of Correct Keypoints (PCK) of all methods for a fair comparison.

In our method, we adopt RegDA as the simple baseline, first trained on the source domain with the same learning rate scheduling as the source only. Then, we built TransPar in RegDA trained with the proposed minimax game for another 30 epochs. Expressly, we simultaneously incorporate TransPar in the feature extractor, the main regressor, and the adversarial regressor. The main regressor and adversarial regressor are both 2-layer convolutional networks with a width of 256. The learning rate of the regressor is set 10 times to that of the feature extractor. We adopt the mini-batch SGD for optimization. Data augmentation strategies include random resized crop, random rotation, color jitter, and Gaussian blur.

\subsubsection{Results}

\begin{table}[ht!]
    \centering
    \caption{PCK (\%) on task \textit{RHD}$\rightarrow$\textit{H3D}. For all kinds of keypoints, our approach outperforms existing domain adaptation considerably.}
    \scalebox{1}{
    \begin{tabular}{c|cccc|c}
        \toprule
        Method
        &MCP & PIP & DIP & Fingertip & Avg \\
        \midrule
        ResNet-101~\cite{DBLP:conf/cvpr/HeZRS16} & 67.4 & 64.2 & 63.3 & 54.8 & 61.8 \\ 
        DAN~\cite{DBLP:conf/icml/LongC0J15} & 59.0 & 57.0 & 56.3 & 48.4 & 55.1 \\ 
        DANN~\cite{DBLP:conf/icml/GaninL15} & 67.3 & 62.6 & 60.9 & 51.2 & 60.6 \\ 
        MCD~\cite{DBLP:conf/cvpr/SaitoWUH18} & 59.1 & 56.1 & 54.7 & 46.9 & 54.6\\
        MDD~\cite{DBLP:conf/icml/0002LLJ19} & 72.7 & 69.6 & 66.2 & 54.4 & 65.2\\
        RegDA~\cite{DBLP:journals/corr/abs-2103-06175} & 78.9 & 72.8 & 70.0 & 61.6 & 70.9 \\ \midrule
        \textbf{TransPar-RegDA} & \textbf{80.8} & \textbf{75.2} & \textbf{70.8} & \textbf{62.0} & \textbf{72.4} \\
        \bottomrule
    \end{tabular}}
    \label{tab:keypoint}
  \end{table}

We use Percentage of Correct Keypoints (PCK) as the evaluation metric. Following~\cite{DBLP:journals/corr/abs-2103-06175}, an estimation is considered correct if its distance from the ground truth is less than a fraction $\alpha$ = 0.05 of the image size. We report the average PCK on all 21 keypoints. We also report PCK at different parts of the hand, such as metacarpophalangeal (MCP), proximal interphalangeal (PIP), distal interphalangeal (DIP), and fingertip.

Table~\ref{tab:keypoint} reports the results on the hand keypoint detection task \textit{RHD}$\rightarrow$\textit{H3D}. We can observe that most existing domain adaptation methods (\eg, DAN, DANN, MCD) perform poorly on the real keypoint detection problem. They even achieve a lower accuracy than source only, and their accuracy on the testing set varies greatly during training. In comparison, TransPar-RegDA has significantly improved the accuracy of ResNet-101 at all positions of hands, and the average accuracy has increased by 10.6\%. Incorporated into RegDA, TransPar-RegDA further improves the accuracy of RegDA by a non-trivial margin and achieves state-of-the-art accuracies at all positions of hands. Specifically, TransPar-RegDA achieves 72.4\% average accuracy, 1.5\% higher than the original RegDA, and 17.3\% higher than the original DAN. The above results explicitly demonstrate that TransPar properly contributes to learning more domain-invariant information and successfully bridges the domain discrepancy in the keypoint detection task, once proving its universality. Since the keypoint detection is a novel domain adaptation regression task, TransPar shows its strong scalability, \ie, it can be easily extended to handle any case of data distribution shift scenarios.

Fig.~\ref{H3D} shows some visualization results before and after adaptation from source only, RegDA, and TransPar-RegDA compared with ground truths. We can observe that the false predictions of source only are often located at the positions of other keypoints, resulting in the predicted skeleton not look like a human hand. In contrast, the outputs of TransPar-RegDA look more like a human hand automatically. In some difficult keypoints, RegDA produces false detection results, while TransPar-RegDA gets robust predictions.

\section{Conclusion and Future Work}
\label{Conclusion}
In this paper, we propose a novel method to distinguish the transferable and untransferable parameters to achieve robust unsupervised domain adaptation. We suggest different update rules for different types of parameters to reduce the memorization of domain-specific information. Extensive experiments on image classification and 2D hand keypoint detection have demonstrated that our method could achieve more accurate performance in various real-world applications. Our approach is simple and orthogonal to other methods. In future work, we believe that our method opens up new possibilities in the topics of unsupervised domain adaptation. One can extend our work into the multi-source domain adaptation, open-set domain adaptation, and other cases of data distribution shift.

\section*{Acknowledgment}
This work is supported by the National Natural Science Foundation of China (61876098), the National Key R\&D Program of China (2018YFC0830100, 2018YFC0830102).

\bibliographystyle{IEEEtran}
\bibliography{aaai}

%

\begin{IEEEbiography}[{\includegraphics[width=1in,height=1.25in,clip,keepaspectratio]{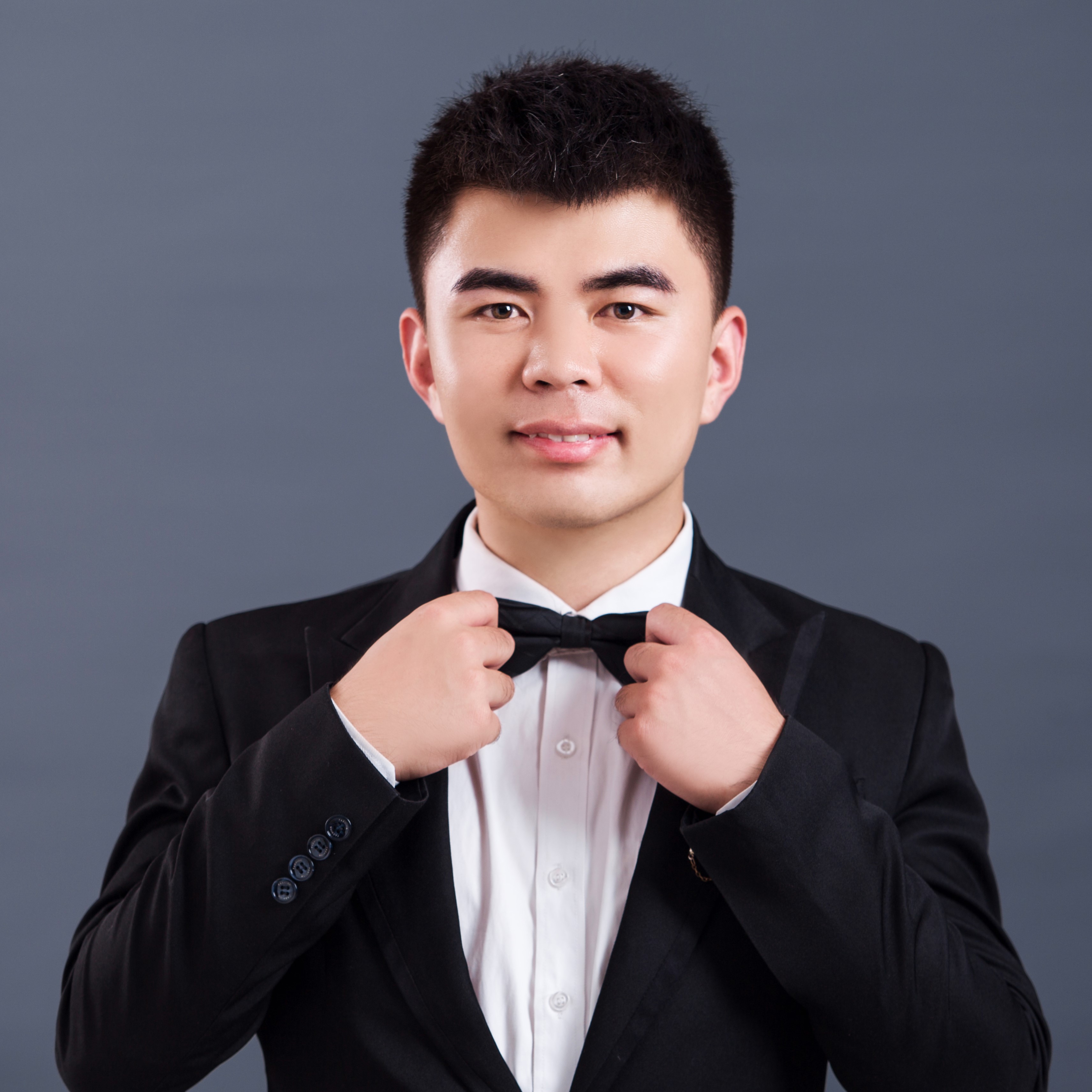}}]{Zhongyi Han} received his B.Eng. and M.M. degree from Shandong University of Traditional Chinese Medicine, in 2016 and 2019, respectively. He had been a visiting student at the University of Western Ontario, CA. Currently, he is working toward a Ph.D. degree with the Artificial Intelligence Research Center in Shandong University, supervised by Prof. Yilong Yin. His research interest is mainly in machine learning, computer vision, and medical image analysis. He has published over 10 papers in leading international journals and conferences.
\end{IEEEbiography}


\begin{IEEEbiography}[{\includegraphics[width=1in,height=1.25in,clip,keepaspectratio]{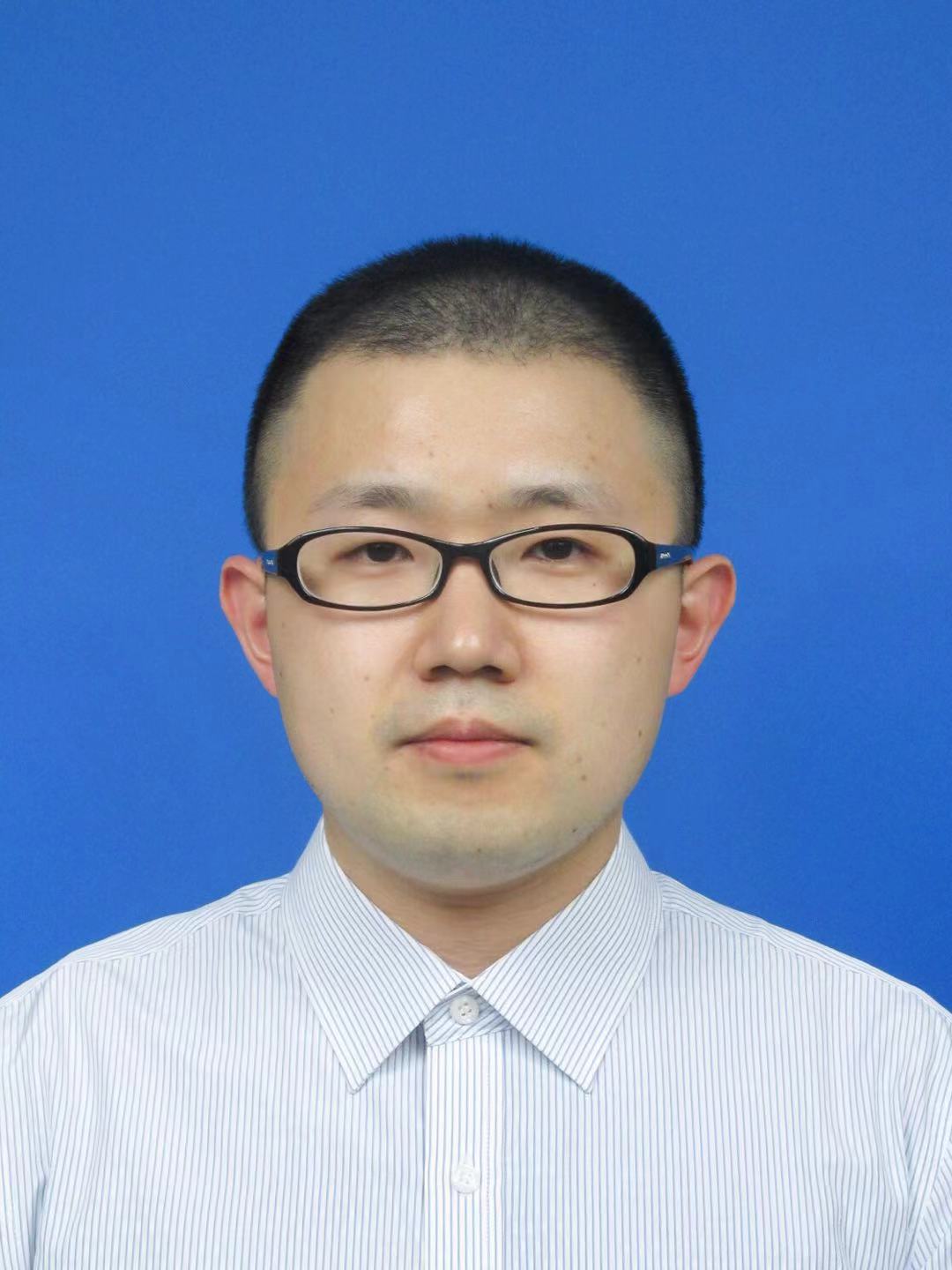}}]{Haoliang Sun} received B.Eng. and Ph.D. from Shandong University in 2014 and 2020, respectively. He had been a visiting student at the University of Western Ontario, CA, and the University of Wisconsin-Madison, USA. He currently pursues the postdoc at Shandong University. His research interests include machine learning, computer vision, and medical image analysis. He has published over 10 papers in leading international journals and conferences.
\end{IEEEbiography}

\begin{IEEEbiography}[{\includegraphics[width=1in,height=1.25in,clip,keepaspectratio]{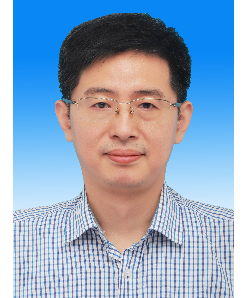}}]{Yilong Yin} received the Ph.D. degree from Jilin University, Changchun, China, in 2000. From 2000 to 2002, he was a Postdoctoral Fellow with the Department of Electronic Science and Engineering, Nanjing University, Nanjing, China. He is the Director of the Artificial Intelligence Research Center and the Professor of Shandong University, Jinan, China. His research interests include machine learning and data mining. He has published over 100 papers in leading international journals and conferences.
\end{IEEEbiography}




\end{document}